\documentclass[lettersize,journal]{IEEEtran}

\usepackage{times}
\usepackage{amsmath,amsfonts,amssymb}
\usepackage{prettyref}
\usepackage{mathrsfs}
\usepackage{graphicx}
\usepackage{wrapfig}

\usepackage{enumitem}
\usepackage{MnSymbol}
\usepackage{multirow}
\usepackage{booktabs}
\usepackage{algorithm}
\usepackage[table]{xcolor}
\usepackage{cite}
\usepackage{ragged2e}
\usepackage{caption}
\usepackage{subcaption}
\usepackage{orcidlink}
\usepackage{graphicx}
\usepackage{float}
\usepackage{rotating}

\usepackage{times}

\usepackage[numbers]{natbib}
\usepackage{multicol}
\usepackage{amsmath}
\usepackage{graphicx}
\usepackage{hyperref}
\usepackage{xcolor}
\usepackage{amsfonts}       
\usepackage{algorithm}
\usepackage[noend]{algpseudocode}
\makeatletter
\newlength{\comment@width}
\usepackage{comment}
\usepackage{auto-pst-pdf} 

\usepackage[numbers]{natbib}
\usepackage{multicol}
\usepackage{amsmath}

\begin{document}

\title{Physics-informed Neural Mapping and Motion Planning in Unknown Environments}

\author{Yuchen Liu\orcidlink{0009-0008-9927-1113}*, Ruiqi Ni\orcidlink{0000-0003-3619-1577}*, and Ahmed H. Qureshi\orcidlink{0000-0003-2104-2333}

\thanks{The authors are with the Department of Computer Science, Purdue University, West Lafayette, IN 47907, USA (e-mail: liu3853@purdue.edu, ni117@purdue.edu, ahqureshi@purdue.edu).}
\thanks{* denotes equal contribution}}
\let\oldtwocolumn\twocolumn
\renewcommand\twocolumn[1][]{%
    \oldtwocolumn[{#1}{
    \begin{center}
    \includegraphics[width=\textwidth,trim=0.0cm 3.4cm 0.0cm 3.4cm,clip]{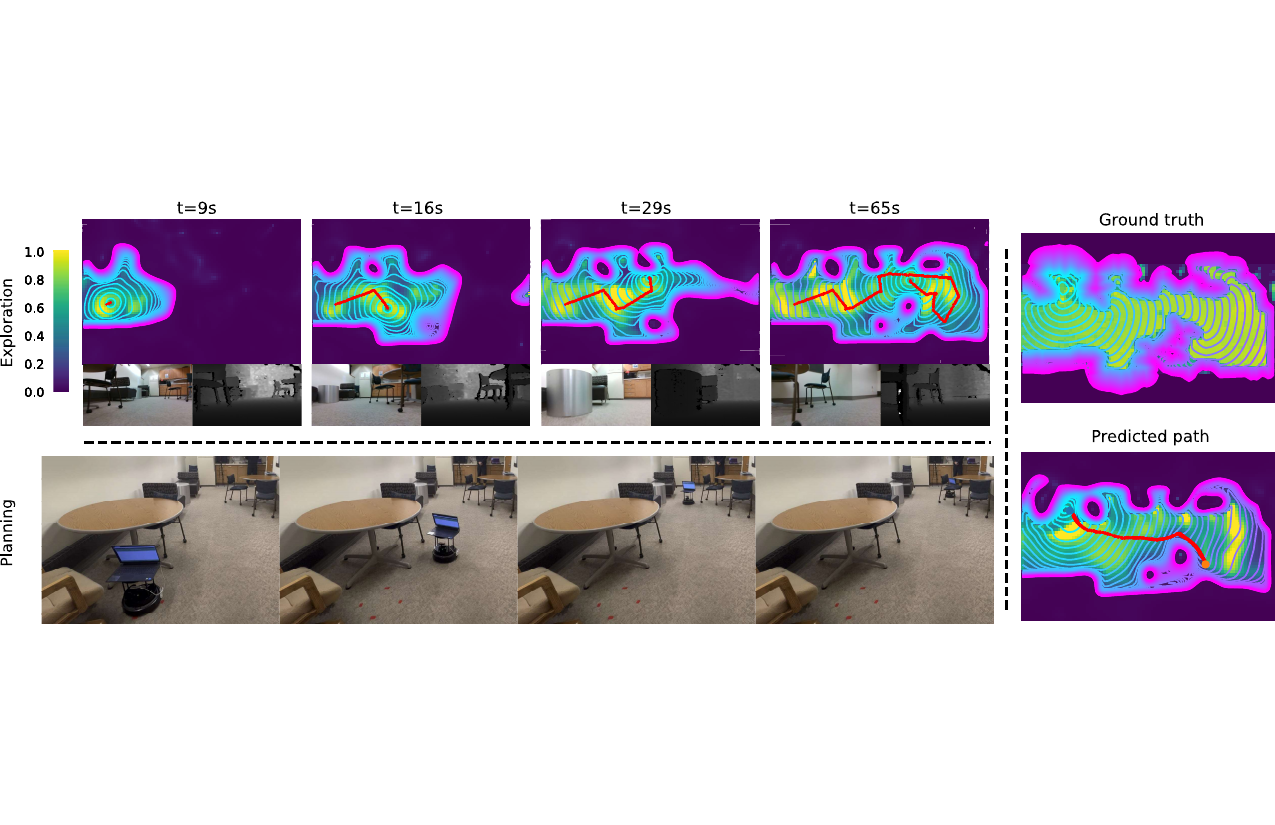}
    \captionof{figure}{\justifying Active NTFields mapping and motion planning in a real-world indoor environment with a differential drive Turtlebot4 robot. \textbf{Mapping:} The top row shows the online exploration of the environment with the corresponding depth image stream. Our method incrementally constructs the map with incoming frames and uses the reconstructed arrival time field in the partially observed environment to reach the next viewpoint for further sensing. Thus, no external motion planner is needed during mapping. Finally, in this environment, the robot takes 65 seconds to actively explore and reconstruct the arrival time fields. \textbf{Motion Planning:} The bottom row shows motion planning in the reconstructed environment, where the goal is set behind the chair, and planning takes only 0.02 seconds.}
           \label{fig:exploration}
        \end{center}
    }]
}
\maketitle
%
\begin{abstract}
Mapping and motion planning are two essential elements of robot intelligence that are interdependent in generating environment maps and navigating around obstacles. The existing mapping methods create maps that require computationally expensive motion planning tools to find a path solution. In this paper, we propose a new mapping feature called arrival time fields, which is a solution to the Eikonal equation. The arrival time fields can directly guide the robot in navigating the given environments. Therefore, this paper introduces a new approach called Active Neural Time Fields (Active NTFields), which is a physics-informed neural framework that actively explores the unknown environment and maps its arrival time field on the fly for robot motion planning. Our method does not require any expert data for learning and uses neural networks to directly solve the Eikonal equation for arrival time field mapping and motion planning. We benchmark our approach against state-of-the-art mapping and motion planning methods and demonstrate its superior performance in both simulated and real-world environments with a differential drive robot and a 6 degrees-of-freedom (DOF) robot manipulator. {The supplementary videos can be found at \url{https://youtu.be/qTPL5a6pRKk}, and the implementation code repository is available at \url{https://github.com/Rtlyc/antfields-demo}}. 
\end{abstract}


\section{Introduction}
{\IEEEPARstart{M}{apping} and motion planning are two fundamental components in robotic operations within unknown environments \cite{choset2005principles,lavalle2006planning}. Mapping is focused on reconstructing environmental features, such as obstacle occupancy or distance to obstacle \cite{choset2005principles,thrun2002probabilistic,stachniss2009robotic}. Motion planning, on the other hand, uses these mapped features to find collision-free sequences of robot movements between the specified start and goal points \cite{lavalle2006planning}. Despite decades of development in both fields \cite{stachniss2009robotic}, a notable gap still exists between effective mapping and efficient motion planning, requiring additional, often computationally intensive, tools to integrate these systems seamlessly.

For instance, a variety of motion planning algorithms—from optimization-based to sampling-based planners—rely heavily on collision features identified during the mapping phase \cite{kuffner2000rrt,zucker2013chomp}. These algorithms, while powerful, demand significant computational resources and time to deliver viable path solutions. Bridging this gap efficiently has profound implications for practical robotics applications.}

{Specifically, robots often need to generate a map of their environment only once and then use it repeatedly to find paths between arbitrary start and goal points. A vacuum robot, for example, may create a map of a home environment only once, but it will use motion planners over that map to reach arbitrary dirty places throughout its lifespan. 
If the mapping feature requires additional computationally expensive tools for motion planning, these tools will be needed repeatedly throughout the robot's operational lifespan. Thus, a map that eliminates the need for costly motion planning methods would be ideal for a wide range of real-world robotics applications.}





{Therefore, in this paper, we propose a novel mapping feature called the arrival time fields, which bridges the gap between mapping and motion planning without needing any computationally expensive tools. 
We show that this new map can be developed online in unknown environments based on local observations. Once the arrival time field map is generated, it can enable motion planning that is significantly faster in computation times than any existing method.}

{The arrival time is determined by solving the Eikonal equation, as described in detail in Section III.} The arrival time represents the shortest travel time from a starting point to a destination. The gradients of the arrival time guide the generation of a continuous shortest path for motion planning \cite{sethian1996fast,ni2023ntfields}. {For example, the Fast Marching Method (FMM) solves the Eikonal equation via grid search over discretized space for path planning \cite{sethian1996fast,valero2013path,treister2016fast}. However, these methods do not scale to higher-dimensional robot configuration space.} Recent methods solve the Eikonal equation via neural networks in continuous space and scale to higher dimensional motion planning problems \citep{ni2023ntfields,Ni-RSS-23,10610883,shen2024pc,li2024riemannian}. Their results show that the arrival time fields allow path planning much faster than any other motion planner. {However, these methods require a known environment and infer arrival time fields through the offline training of neural networks. Additionally, their process to solve the Eikonal equation is either slow or prone to errors in complex environments, making them unsuitable for online applications. For more details on these previous neural Eikonal equation solvers, please refer to Section III. Experimental comparisons with previous methods are provided in Section V.

In contrast to these earlier methods that require a known environment and a computationally intensive training process, the solution to the mapping problem demands a fast, efficient, and reliable methodology for generating a map of an unknown environment. Therefore, the objective of this work is to solve the Eikonal equation using a neural network in unknown environments based on local observations without relying on costly training strategies and complex loss functions.
} 

To achieve the above-mentioned objectives, this paper introduces a novel approach called Active Neural Time Fields (Active NTFields), which quickly maps the arrival time fields in unknown environments in an online manner and uses them for real-time motion planning in complex environments. The main contributions of our proposed work are listed as follows: 
\begin{itemize}[leftmargin=*]
    \item {A new mapping feature called arrival time fields, which allows fast motion planning and scales to higher dimensional robot configuration space.}
    
    \item A novel framework that actively explores the unknown environment and uses only local perception data to train neural networks on the fly to map the arrival time field of the given, unknown environment. 
       
    \item A fast motion planning method that does not require any complex tools and instead directly uses the gradients of the arrival time field map for pathfinding in near real-time.
    \item Demonstrations of our proposed mapping and motion planning framework in simulated and real-world, complex environments with a differential drive robot and 6-DOF robot manipulator.
\end{itemize}
Our results show that the Active NTFields quickly map the unknown environment and allow motion planning much faster than any existing method. Furthermore, it also scales to real-world settings, including kitchen and narrow-passage cabinet-like environments.

\section{Related Work}

\noindent Mapping the environment encodes the basic kinematic constraints for motion planning. Commonly, these constraints are used for collision avoidance, which dominates the computational cost in motion planning \citep{bialkowski2011massively}. Map representations could be obstacle representations like point clouds. They are surface geometry, which is easily gathered from the sensor's back-projected rays \citep{Ortiz-RSS-22}. However, they are impractical for motion planning due to collision querying with logarithmic computational times of map size \citep{pan2010gpu, bialkowski2016efficient}, especially for raw real-world data. Signed Distance Field (SDF) and truncated SDF provide an alternative map representation for collision avoidance to compute distance in constant time. They can be quickly constructed by fusing depth measurements \citep{newcombe2011kinectfusion,newcombe2015dynamicfusion}. These methods are grid-based and are limited to a large resolution due to computational and memory costs \citep{han2019fiesta,oleynikova2017voxblox}.

Recently, neural fields have emerged as suitable environment representations due to their compactness and continuity property \citep{park2019deepsdf,mescheder2019occupancy,chen2019learning}. They use a neural network to map a coordinate to some signal like occupancy or distance \citep{xie2022neural}, and they can be trained from scratch to accurately fit a specific scene \citep{azinovic2022neural,sitzmann2019siren,yan2021continual}. 
Besides offline reconstruction tasks, neural fields can also be trained in real-time as part of a SLAM system \citep{sucar2021imap,Zhu2021NICESLAMNI,Ortiz-RSS-22,sandstrom2023point}. For instance, iMAP \citep{sucar2021imap} was the first work to reconstruct the map by real-time continual learning of neural radiance fields. Similarly, iSDF \citep{Ortiz-RSS-22} reconstructed the signed distance field in unknown environments.

While the SDF excels in performing efficient batch collision queries, finding paths within the environment still requires the incorporation of motion planning methods such as sampling-based methods (SMP) \citep{kavraki1998analysis,bohlin2000path, kuffner2000rrt} and trajectory optimization (TO) \citep{von1992direct, betts1993path}. SMP still suffer from significant computation times in high-dimensional space, and they struggle to find optimal, even valid results due to low sampling efficiency \citep{qureshi2016potential, tahir2018potentially, gammell2014informed}. In contrast, iterative optimization in TO relies on initial trajectories, but the solutions are not global trajectories and frequently converge to local minima \citep{zucker2013chomp,schulman2014motion,Ni2021RobustMT,le2023accelerating}. This issue persists in neural environment representations, such as neural SDF \citep{adamkiewicz2022vision, camps2022learning}. 

{One exception is roadmap-based SMPs \cite{kavraki1996probabilistic,bohlin2000path}, which share the same philosophy of building a map suitable for motion planning. For example, PRM can be seen as a mapping method that maps the collision-free C-Space by graph construction. Recent work also shows its effectiveness in constructing collision-free graphs in unknown environments through exploration \citep{dang2020graph,xu2021autonomous,yang2022far}. However, the explicit graph representation is discrete, which means that the start and goal positions are approximated to the nearest nodes in the graph, and it can be computationally demanding to scale to high-dimensional configuration spaces. Additionally, motion planning on the constructed graph relies on graph search algorithms and often requires additional trajectory optimizers to smooth the path.}

In recent times, Neural Motion Planners (NMP) ~\citep{qureshi2019motion,qureshi2020motion,ichter2018learning, qureshi2018deeply, kumar2019lego, fishman2023motion} based on learning have emerged. They prove efficient in finding paths, especially with prior knowledge, and can scale to high DOF robot systems. These planners take raw environmental data, such as point clouds or depth images, as input and forecast trajectories. However, they are limited by their need for offline training because their training data relies on expert trajectories from traditional planners. Consequently, these methods are unsuitable for real-time mapping scenarios where experts are not in the loop.

Perhaps the closest relevant mapping method is the cost-to-go map \citep{chen2016motion,tamar2016value,chaplot2021differentiable,huh2021cost, li2022learning}. This map stores the cost-to-go function in C-Space coordinates, and the gradients of this function guide motion planning paths. Traditionally, the cost-to-go function relies on discrete representations of free space environments, calculated through wavefront \citep{sethian1996fast,valero2013path,treister2016fast} or diffusion methods \citep{connolly1990path,crane2013geodesics,chen2016motion, crane2020survey}. However, these methods are inefficient and challenging to scale for high DOF conditions.
Learning-based approaches have been suggested to learn the cost-to-go function on grid \cite{tamar2016value,chaplot2021differentiable} or continuous space \citep{huh2021cost, li2022learning}. Unfortunately, they necessitate expert cost-to-go functions as supervision data, leading to time-consuming data generation and offline training. 

Recently, physics-informed neural networks (PINN) \cite{raissi2019physics} have emerged as promising tools for motion planning. They find path solutions in a fraction of a second and do not require expensive expert trajectories for learning. NTFields \citep{ni2023ntfields} and its variant P-NTFields \citep{Ni-RSS-23} are the first of these methods that directly learn to solve the Eikonal equation without needing expert trajectories. The solution of the Eikonal equation is an arrival time field between the given start and goal. The arrival time field is a type of cost-to-go function whose gradients lead to a path solution. The NTFields were limited by the complex loss landscape of the Eikonal equation. The P-NTFields extended the Eikonal equation via Laplacian and introduced progressive speed scheduling to guide neural network learning. However, these additions significantly increased the training time of P-NTFields, making it unsuitable for online mapping tasks of unknown environments.

In this paper, we propose an innovative framework named Active NTFields, incorporating a new formulation of the Eikonal equation alongside a novel neural architecture and a loss function without the need for Laplacian or curriculum learning. Our framework facilitates rapid active learning and efficient mapping of neural time fields within unknown environments. Once learned, these time fields allow pathfinding in a negligible amount of time, much faster than state-of-the-art motion planners. 

\section{Background}
\noindent Let the environment be denoted as $\mathcal{X}\in \mathbb{R}^3$ with its obstacle and obstacle-free space as $\mathcal{X}_{obs}$ and $\mathcal{X}_{free}$, respectively. The robot configuration space is indicated as $\mathcal{Q} \in \mathbb{R}^d$ with dimension $d$, where $\mathcal{Q}_{obs}$ and $\mathcal{Q}_{free}$ indicate the obstacle and obstacle-free space. The objective of environment mapping is for a robot to explore environment $\mathcal{X}$ to recover its features $F$ \cite{thrun2002probabilistic,stachniss2009robotic}. The traditional mapping methods describe the features as maps, indicating obstacles and obstacle-free space \cite{stachniss2009robotic}. The modern approaches also recover the environment's geometry as the SDF \cite{Ortiz-RSS-22,sandstrom2023point}. The SDF provides the signed distance of any point in the environment to its obstacle geometry's surface. These features allow motion planning methods to find a collision-free path connecting a robot's given start and goal configurations to navigate the mapped environment \cite{zucker2013chomp}. The collision avoidance constraints for the robot path are satisfied by leveraging the environment features $F$ recovered during mapping. However, traditional and modern mapping features require an extra computational mechanism to find the robot's motion path. In this paper, we introduce a new mapping feature called the arrival time field, which is the solution to the Eikonal equation. 

{The Eikonal equation is a first-order non-linear equation that represents a robot moving from a start $q_s$ to a goal $q_g$ within a speed constraint $S(q)$ defined over robot configurations. The speed function outputs a scalar value and is designed so that the robot's speed is high in the free space and low near the obstacle region. The Eikonal equation relates the speed constraint to the robot arrival time $T$ between the start $q_s$ and goal $q_g$,  i.e.,}
\begin{equation}
\frac{1}{S(q_g)} = \|\nabla_{q_g} T(q_s,q_g)\|
\label{eikonal}
\end{equation}
The solution of the above PDE is the shortest arrival time between the given start and goal. The prior methods solved the above equation using the grid-based FMM to recover the shortest path based on arrival time between the given start and goal \citep{sethian1996fast,valero2013path,treister2016fast}. However, these methods lacked continuity and were computationally intractable in higher dimensional spaces. Recent work introduced a Laplacian-based viscosity term in the Eikonal equation and a new factorized formulation of arrival time, which allowed for solving Eq. \ref{eikonal} using neural networks \citep{ni2023ntfields,Ni-RSS-23}. The factorized arrival time is described as follows:
\begin{equation}
T(q_s,q_g)=\cfrac{\|q_s-q_g\|}{\tau(q_s,q_g)}   
\label{factorized}
\end{equation}
The neural network predicts $\tau$ for the given start and goal to parameterize the above equation, which leads to arrival time $T$ and its gradient. When the target point is in the obstacle space, the $\tau$ is predicted to be 0, making $T$ very large, and when the target is the same as the start, $\|q_s-q_g\|$ is 0, making arrival time $T$ equal to zero. The estimated arrival time $T$ and its gradient are then used to predict the speed using the following viscosity Eikonal equation with a Laplacian \citep{crandall1983viscosity}.
\begin{equation}
\frac{1}{S(q_g)} = \|\nabla_{q_g} T(q_s,q_g)\| + \epsilon \Delta_{q_g} T(q_s,q_g),
\label{viscoeikonal}
\vspace{-0.05in}\end{equation}
where $\epsilon$ is a hyperparameter. The above formulation is semi-linear elliptic and has a unique solution around 
obstacles. Finally, the predicted speed is compared against the following ground truth speed function $S^*$ via isotropic loss function to train the neural network.
\begin{equation}
S^*(q)=\cfrac{s_{const}}{d_{max}}\times\mathrm{clip}(\boldsymbol{\mathrm{d}}(q,\mathcal{X}_{obs}), d_{min}, d_{max})  
\label{speed}
\end{equation}
The function $\boldsymbol{\mathrm{d}}$ provides the minimum distance from robot points of C-Space point $q$ to the obstacle $\mathcal{X}_{obs}$, clipped between user-defined thresholds $d_{min}$ and $d_{max}$, and scaled by factor $s_{const}$. {Note that the output of $S^*$ is a scalar value and is based on the minimum distance of robot configuration to the obstacle.} Once the neural network is trained, it provides the arrival time field between any given start and goal, leading to a path solution. For further details, refer to \citep{ni2023ntfields,Ni-RSS-23}.

\section{Proposed Method}

\begin{figure}[!t]
    \centering
        \centering
        \includegraphics[width=0.49\textwidth,trim=0.0cm 1.5cm 0.0cm 1.5cm,clip]{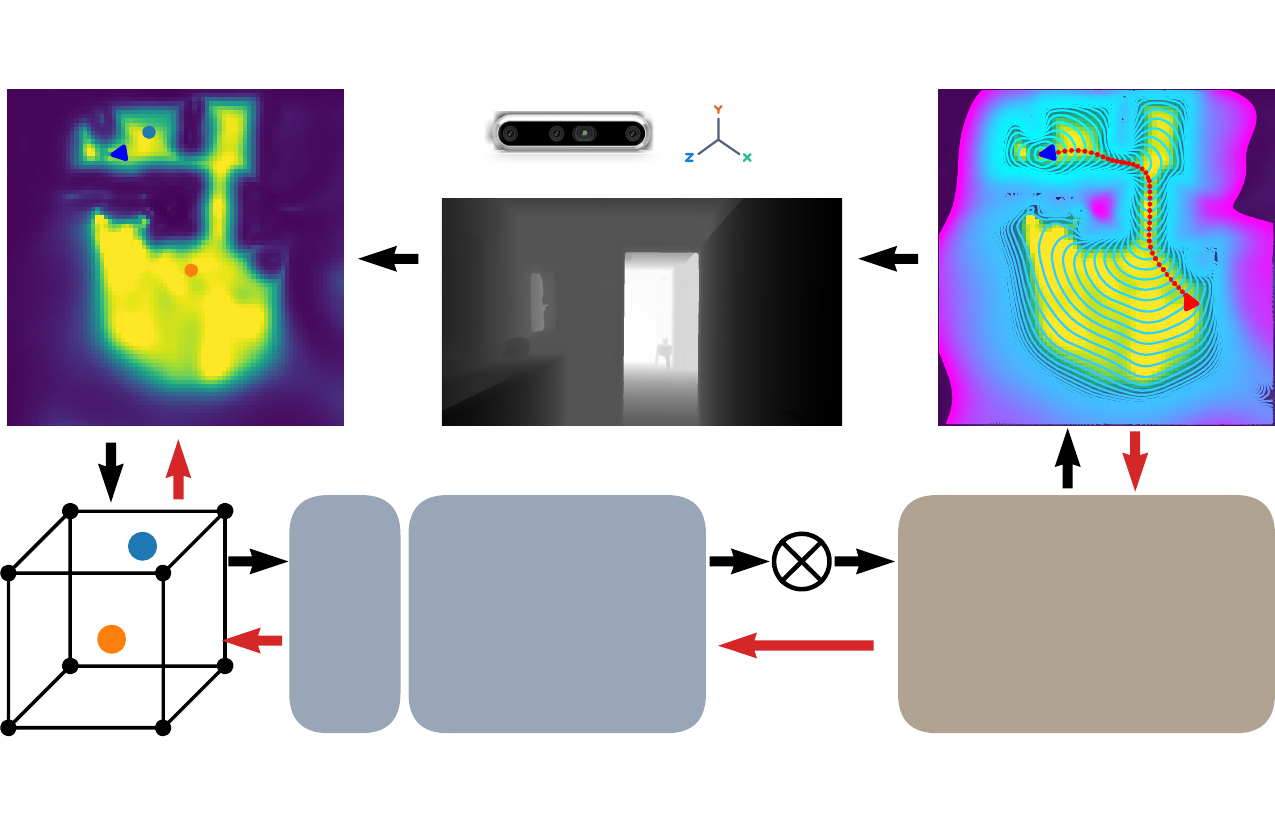}
    \put(-214,52.2){{{ {\footnotesize\begin{math}1/S=\|\nabla T\|\end{math}}}}}
    \put(-98.5,10){{ {\footnotesize$\nabla T$}}}
    \put(-52,52.2){{\color{black}{\footnotesize$ T$}}}
    %
    \put(-52,130){{\color{black}{\footnotesize Time Field}}}
    \put(-236,130){{\color{black}{\footnotesize Speed Field}}}
    \put(-144,130){{\color{black}{\footnotesize Posed Sensor}}}
    \put(-220.7,40){{\footnotesize$q_s$}}
    \put(-227.7,20.5){{\footnotesize$q_g$}}
    \put(-191,24){{{\footnotesize\color{black}{\footnotesize$\gamma(\cdot)$}}}}
    \put(-154.5,31){{\footnotesize C-Space }}
    \put(-154.5,24){{\footnotesize  Encoder}}
    \put(-148,15){{{\footnotesize\color{black}$f(\cdot)$}}}
    \put(-54.5,31){{\footnotesize Time Field}}
    \put(-52.5,24){{\footnotesize Generator}}
    \put(-44,15){{{\footnotesize\color{black}$g(\cdot)$}}}
    %
    \caption{ \justifying 
    { The process begins by capturing the scene from the camera's current viewpoint and sampling C-space points \( q_s \) and \( q_g \). Each sampled point pair \( (q_s, q_g) \) is then input into the neural network, as shown in Fig. \ref{fig:structure}, to predict the corresponding arrival times \( T \). The predicted speed field is derived using the Eikonal equation, as detailed in Eq. \ref{predictspeed}. The neural network is trained based on the loss between the ground truth speed and the predicted speed, as outlined in Eq. \ref{trainloss}. Finally, the next viewpoint is selected using an exploration strategy, and the local arrival time fields guide the robot to that viewpoint without requiring any external motion planner. This process repeats until the entire environment is observed and its arrival time fields are reconstructed.} 
    }
    \label{fig:pipeline}
    \vspace{-0.2in}
\end{figure}

\noindent This section presents our proposed framework for actively mapping the arrival time field of the environment and utilizing it for path planning. The key components and their integration into a unified Active NTField framework are described as follows. 
\subsection{New Time Field Factorization}
Although the factorization in Eq. \ref{factorized} has the desirable properties for motion planning, the arrival time $T$ increases steeply as $\tau$ decreases to zero. Hence, when computing the gradients of Eq. \ref{eikonal} to train the neural network, the sharp features around 
obstacles often lead to incorrect local minima. Therefore, in P-NTFields \citep{Ni-RSS-23}, a Laplacian term in Eq. \ref{viscoeikonal} and a speed scheduler was introduced. The former leads to a unique Eikonal equation solution, whereas the latter progressively reduces the speed around obstacles as the training epoch increases to prevent convergence to an incorrect solution. The Laplacian is computationally expensive to calculate as it requires the second derivation of the neural network, and the progressive speed scheduling requires a large number of training epochs. Both strategies make P-NTFields unsuitable when aiming for a fast mapping approach that learns to infer the arrival time field of the environment on the fly. Hence, we introduce a new factorization of the arrival time as described in the following. 
\begin{equation}
T(q_s,q_g)=\log(\tau(q_s,q_g))^2\|q_s-q_g\|   
\label{factorized2}
\end{equation}
It can be seen that we replace $1/{\tau}$ in Eq. \ref{factorized} with ${\log(\tau)^2}$. The latter is more flattened and changes less steeply as $\tau$ decreases, preventing sharp features near obstacles. We show in our experiments that the new definition recovers accurate arrival time fields without needing Laplacian or speed scheduling. Furthermore, by solving the Eikonal equation (Eq. \ref{eikonal}) via chain rule with the new arrival time definition in Eq. \ref{factorized2}, the speed $S$ becomes as follows:
\begin{equation}
{S(q_g)=1/\|\nabla_{q_g}(\log(\tau(q_s,q_g))^2\|q_s-q_g\|) \|}
\label{predictspeed}
\end{equation}
\subsection{Neural Model Design} Our neural architecture is inspired by spectral distance (SD) formulation \citep{coifman2006diffusion,lipman2010biharmonic}. The SD $d_w(q_s, q_g)$ between two points $q_s$ and $q_g$ is an approximate alternative to an arrival time with the following form:
\begin{equation}
d_w(q_s, q_g)^2 = \sum_{i=1}^{\infty} w(\lambda_i)(\phi_i(q_s)-\phi_i(q_g))^2,
\label{spectral}
\end{equation}
where $\phi_i(\cdot)$ and $\lambda_i$ are the eigenfunctions and eigenvalues, respectively, of the Laplace operator {$\Delta$} in the given collision-free space and $w(\cdot)$ is a weight function. {The eigenfunction of the Laplace operator in the
given collision-free space implies $-\Delta_q\phi_i(q)=\lambda_i\phi_i(q)$ when $
q\in\mathcal{Q}_{free}$, {whereas $\nabla_\mathbf{n} \phi_i(q) = 0$ when $ q\in\partial\mathcal{Q}_{free}$ and $\nabla_\mathbf{n}$ denotes the directional derivative along the outward normal of the collision-free space boundary.}} 


In practice, the spectral distance is computed using the $k$ smallest eigenvalues and their corresponding eigenfunctions. { Although SD does not produce the exact shortest arrival time, it provides a robust global approximation sufficient to retrieve a path solution, as illustrated in Fig. \ref{fig:perfcompare}. This attribute of SD to capture global structure inspires our neural architecture design to capture the global structure of the time field.}

This SD formulation inspires our neural time field network to approximate the shortest arrival time. To further improve its representational capacity, we integrate random Fourier positional encoding \citep{tancik2020fourier} with the SIREN neural network architecture \citep{sitzmann2019siren}.



Given the robot start $q_s \in \mathcal{Q}_{free}$ and goal  $q_g \in \mathcal{Q}_{free}$, we obtain their Fourier positional encoding as follows:
\begin{equation}
\begin{aligned}
    &\gamma(q_s)=[\cos(2\pi \mathbf{c}^T q_s),\sin(2\pi \mathbf{c}^T q_s)]\\ &\gamma(q_g)=[\cos(2\pi \mathbf{c}^T q_g),\sin(2\pi \mathbf{c}^T q_g)],
    \end{aligned}
    \label{rff}
\end{equation}
where $\mathbf{c}$ is a fixed latent random code. These encodings are then passed through the SIREN neural network, denoted as $f$, which is a multilayer perceptron with sine activation functions. The sine activation function captures the high-frequency features and allows smooth differentiation for gradient computation in our method. Let the combination of positional and SIREN be defined as $\Phi(q)=f(\gamma(q))$. Next, as inspired by Eq. \ref{spectral}, we subtract these features and take a square, i.e., 
\begin{equation}
{
\begin{aligned}
    &{\Phi(q_s)\bigotimes \Phi(q_g)=(\Phi(q_s)-\Phi(q_g))^2 \text{, where}} \\ &\Phi(q_s)=f(\gamma(q_s)), \Phi(q_g)=f(\gamma(q_g))
    \end{aligned}}
\label{operator}
\end{equation}

{The above squared subtraction is also relevant to the symmetric operator introduced in NTFields \citep{ni2023ntfields}. This operator enforces the arrival time field's symmetric property, i.e., the arrival time from start to goal and from goal to start should be the same. 
Specifically, let latent encodings of start and goal configurations be denoted as $a$ and $b$. The symmetric operator in \citep{ni2023ntfields} was defined as the concatenation of min and max, i.e., $[\max(a,b), \min(a,b)]$. The other choice of symmetric operators could be subtraction, addition, or averaging. 
Hence, the squared subtraction operator in Eq. \ref{operator}, inspired by spectral distance, also acts as a symmetric operator, ensuring the symmetric property of the arrival time field. Furthermore, we observed that the squared subtraction operator instead of concatenation of min and max reduces the feature size, leading to neural network training efficiency. Finally, the resulting features from Eq. \ref{operator} are given to another multi-layer perceptron, denoted as $g$, that predicts the $\tau$.} In summary, our neural time field network with parameters $\theta$ is summarized as follows:

\begin{equation}
    \tau_\theta(q_s,q_g)=g\big(\Phi(q_s)\bigotimes \Phi(q_g)\big)
\label{tf}\end{equation}

As highlighted in Eqs. \ref{operator}-\ref{tf} and Fig. \ref{fig:pipeline}, our model has four parts, i.e., the positional encoder $\gamma(\cdot)$, configuration space (C-Space) encoder $f(\cdot)$, symmetric operator $\bigotimes$, and time field generator $g(\cdot)$. The positional encoder, $\gamma (q)$, takes a given configuration $q$ and outputs the Fourier features, according to Eq. \ref{rff}. In our implementation, these features are of size 256 units. These features are then passed through function $f(\cdot)$, which comprises four fully connected (FC) layers followed by Sine activation to produce C-Space features, i.e., $f(\gamma(q))$. The number of hidden units of FC layers in function $f(\cdot)$ is 128, as also indicated in Fig. \ref{fig:structure} (a). By following the above-mentioned procedure, we obtain the C-space features of both start and goal configurations, i.e., $f(\gamma(q_s))$ and $f(\gamma(q_g))$. Next, the Symmetric Operator $\bigotimes$ preserves the invariance and symmetry property when switching the start and goal configurations by subtracting and squaring the C-space encodings (Fig. \ref{fig:structure} (b)). Finally, the time field function $g(\cdot)$ takes the symmetrized latent embedding and processes them through multiple FC and Softplus blocks to output a Sigmoid value representing the factorized time field. The number of hidden units of FC layers in function $g (\cdot)$ is 128 (Fig. \ref{fig:structure} (b)).

\begin{figure}[!ht]
\centering
\includegraphics[width=0.45\textwidth,trim=2.4cm 0.0cm 2.4cm 0.0cm,clip]{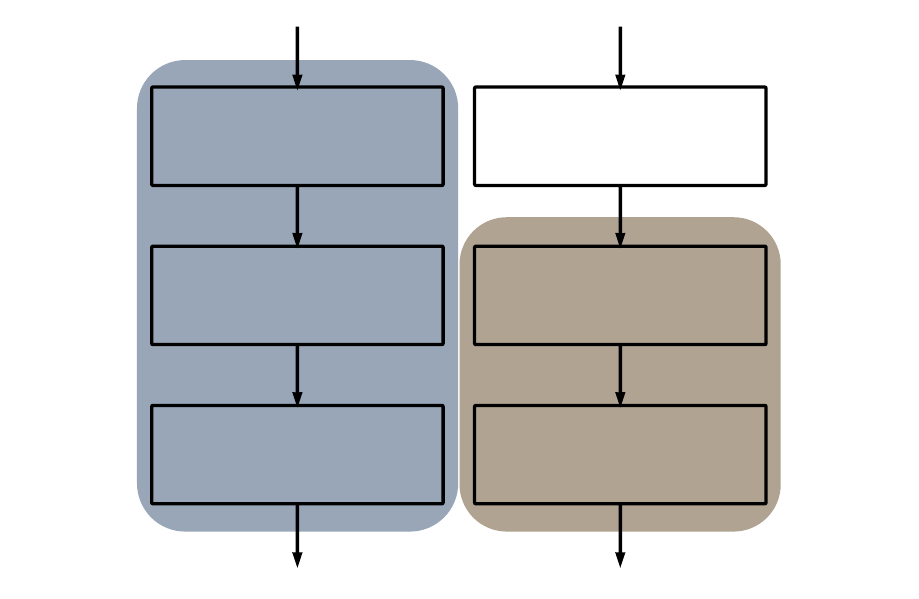} 
\footnotesize{
\put(-178,215){$q$}
\put(-93,215){$f(\gamma(q_s)), f(\gamma(q_g))$}
\put(-214,178){Positional encoding $\gamma(q)$}
\put(-222,160){$[\cos(2\pi \mathbf{c}^T q),\sin(2\pi \mathbf{c}^T q)]$}
\put(-195,114){FC + Sine}
\put(-197,100){$256\times128$}
\put(-205,55){(FC + Sine) $\times$ 3}
\put(-197,41){$128\times128$}
\put(-216,0){C-Space feature $f(\gamma(q))$}
%
\put(-95,178){Symmetric Operator $\bigotimes$}
\put(-100,160){$(f(\gamma(q_s))-f(\gamma(q_g)))^2$}
\put(-97,114){(FC + Softplus) $\times$ 3}
\put(-75,100){$128\times128$}
\put(-80,55){FC + Sigmoid}
\put(-75,41){$128\times1$}
\put(-80,0){Time Field $g(\cdot)$}
%
}
\normalsize{
\put(-222,-25){(a) C-Space Encoder}
\put(-107,-25){(b) Time Field Generator}
}
\caption{{ Our Neural Architecture: The fully connected (FC) layer, with the shape depicted below, forms the core of our network. We employ several activation functions, including Sine, Softplus, and Sigmoid. In (a), $q$ represents a point in the configuration space, $\gamma(\cdot)$ denotes a positional embedding, and $f(\cdot)$ (shown in gray) represents a C-space embedding, as also depicted in Fig. \ref{fig:pipeline}. In (b), $g(\cdot)$ (shown in brown) represents the factorized time field, which also appears in Fig. \ref{fig:pipeline}. \label{fig:structure}
\label{fig:structure}}}
\end{figure}
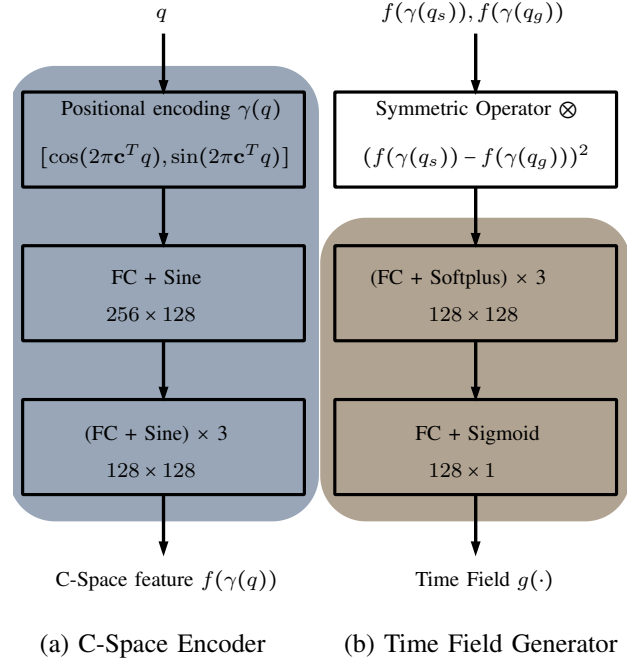


\subsection{Speed Inference}
The neural framework mentioned above predicts $\tau(q_s,q_g)$ for the given start and goal pairs. We use the predicted $\tau$ and its gradients with respect to start $\nabla_{q_s}\tau(q_s,q_g)$ and goal $\nabla_{q_g}\tau(q_s,q_g)$ to estimate the speeds $S(q_s)$ and $S(q_g)$ according to Eq. \ref{predictspeed}. These predicted speeds at the start and goal are then utilized to train the network by comparing them against the expert speed model, as described in {Section III}. 
\subsection{Path Inference}
Once the neural time field network is trained, we utilize its predictions and gradients to recover the path solution between any start and goal as follows. First, the network predicts $\tau (q_s,q_g)$, and we use it to predict speeds ({Section IV-A}) and also to parameterize Eq. \ref{factorized2} to determine the arrival time field $T(q_s,q_g)$. Next, we recover the path solution in the same way as in NTFields \citep{ni2023ntfields}, i.e., we follow gradients bidirectionally from start to goal and from goal to start as described below:   
\begin{equation}
    \begin{aligned}
    q_s &\gets q_s-\alpha S^2(q_s)\nabla_{q_s} T(q_s,q_g) \\
    q_g &\gets q_g-\alpha S^2(q_g)\nabla_{q_g} T(q_s,q_g)
    \end{aligned}
    \label{plan}
\end{equation}
The $\alpha$ is a user-defined step-size scaling factor. The above iterative procedure is repeated until the waypoints from the start and the goal are within the threshold. {It should be emphasized that gradient descent does not ensure convergence, and the algorithm may terminate unsuccessfully after a predefined number of iterations.} { To address this issue, the learned arrival time field maps can also be used more robustly by incorporating them into search \cite{jaillet2010sampling} or optimization \cite{williams2016aggressive,bharadhwaj2020model} techniques. These methods will guided by the learned arrival time field map, yielding effective path solutions fast.} 

\subsection{Active Neural Time Fields}
This section describes our approach to gathering data online in unknown environments for training our neural network on the fly. The prior work assumes the environment to be known, and therefore, the dataset for training the neural time field models is gathered offline. The dataset comprises randomly sampled start and goal points and their ground truth speed values. As described in Section IV-A (Eq. \ref{speed}), the ground truth speed $S^*$ is computed for each sampled point based on its minimum distance from the obstacles. The NTFields \citep{ni2023ntfields} and P-NTFields \citep{Ni-RSS-23} can directly acquire this distance from a known environment by calculating the distance between the point and the nearest obstacle mesh. However, in our setting, the environment is unknown, and the robot needs to explore, obtain data, and determine its ground truth speed values to train the neural networks. Therefore, this section provides procedures to actively create such data and define their ground truth speed for online training of our neural networks. 

\subsubsection{Local Perception Processing}\label{e-1}
Our local perception processing is inspired by the iSDF framework \citep{Ortiz-RSS-22}. The raw input of data comprises robot odometry and the sensor readings. We assume the sensor data to be either lidar-based scanning or depth images. The lidar provides scanning in the form of rays. If the perception is from depth images, we sample pixels and convert them to rays using the camera intrinsic matrix. Furthermore, we transform these rays to world coordinates and randomly select $N$ number of rays from the given scan. For each ray, we sample $m \in \mathbb{N}$ stratified points along its range. This means that each ray of length $l \in \mathbb{R}$ will be split into $l/m$ bins, and in each bin, we will select one sample. Hence, we gather $n \in \mathbb{N}$ sample points, denoted as $\boldsymbol{\mathrm{x}}_{\{n\}} \subset \mathcal{X}$, from $N$ rays.  

Next, we need to determine the ground truth speed values for all sampled points. In our setting, the actual minimum distance to the nearest obstacle is unknown since the environment will only be partially observed during exploration. This contrasts with prior methods where the environment mesh is provided, and the nearest distance to obstacles is readily available. Therefore, in our framework, we revert to approximating the minimum distance of sampled points to the obstacles. We calculate the distance between the point and the nearest surface point in the given scan. Although using all lidar points or depth points can be more accurate for distance estimation, it is computationally expensive. Therefore, we only use $N$ sampled rays to select the nearest obstacle surface. For 3D point robot, workspace sampled points $\boldsymbol{\mathrm{x}}_{\{n\}}$ are the C-Space points $\boldsymbol{\mathrm{q}}_{\{n\}}$. However, in the case of a higher-dimension manipulator, we need a further operation to get  C-Space samples and their expert speed values. Specifically, we use $\boldsymbol{\mathrm{x}}_{\{n\}}$ as the manipulator end effect positions and determine their configuration samples $\boldsymbol{\mathrm{q}}_{\{n\}}$ via inverse kinematic solution. Next, for these C-Space samples, we determine the robot surface points to compute the minimum distance to the obstacle. Once the approximate minimum distance of all sampled points from the obstacle is available, we define the expert speed model based on Eq. \ref{speed} (Section IV-A). In summary, this module processes local perception, creating sample points $\boldsymbol{\mathrm{q}}_{\{n\}}$ and their expert speed values $S^*_{\{n\}}$. 

\subsubsection{Online Training} \label{e-2}
Note that our data arrives in streams. Therefore, we maintain a memory buffer, $\mathcal{M}$, that stores all sample points and their ground truth speed gathered over time. However, training the timefield neural network on the complete memory is computationally expensive. Therefore, we randomly sample a batch of points and their ground truth speed from the memory. Let the batch be denoted as $\mathcal{B} \subset \mathcal{M}$, comprising sample points and their ground truth speed values. Next, we augment the memory buffer, $\mathcal{M}=\mathcal{M} \cup (\boldsymbol{\mathrm{q}}_{\{n\}},S^*_{\{n\}})$, and the batch buffer, $\mathcal{B}=\mathcal{B} \cup (\boldsymbol{\mathrm{q}}_{\{n\}},S^*_{\{n\}})$ , with new data. We randomly shuffle the batch buffer and form pairs of consecutive points. These pairs act as start and goal points for training our neural model. The neural network trains on these pairs for a fixed number of iterations. Note that buffer $\mathcal{B}$ contains samples from new and old perception data, and random shuffling allows start and goal pairs to spread across different parts of the observed environment. Hence, it allows neural networks to learn to generate the arrival time field of a fully observed environment over time. 

We pass the sampled start and goal pairs through our neural time field network to obtain the $\tau$. The $\tau$ and its gradients parametrize Eq. \ref{predictspeed} to infer the speed $S$ ({Section IV-A}). Next, we compute the loss of predicted speed against ground truth speed and use it to train our neural networks for a fixed number of iterations. 
\begin{equation}
    \begin{aligned}
L(S^*,S)=    &(\sqrt{S^*(q_s)/S(q_s)}-1)^2+\\
    &(\sqrt{S^*(q_g)/S(q_g)}-1)^2
    \end{aligned}
    \label{trainloss}
\end{equation}
Note that the above loss function is different from the isotropic loss function used in prior work \citep{ni2023ntfields,Ni-RSS-23}. Since we proposed a more flattened version of factorized time, we expect the loss to be also flattened. Therefore, our new loss with squared root and square makes the loss smoother and more flat than MSE and isotropic loss functions. We observed that our new loss function aids in training our neural network faster than prior loss functions introduced by NTFields \citep{ni2023ntfields} and P-NTFields \citep{Ni-RSS-23}.

\begin{algorithm}
    \caption{Active NTFields Online Learning}
    \label{alg:active_exploration}
    \begin{algorithmic}[1]
        \Require \\
        \begin{itemize}
            \item $\text{Odom data}$ \hfill \Comment{Robot odometry data}
            \item $\text{Lidar or depth images}$ \hfill \Comment{Sensor readings}
            \item $N$ \hfill \Comment{Number of rays}
            \item $m$ \hfill \Comment{Number of stratified points per ray}
            \item $\mathcal{M}$ \hfill \Comment{Memory buffer}
            \item $\mathcal{B}$ \hfill \Comment{Batch buffer}
            \item $\theta$ \hfill \Comment{Model Parameters}
            \item $\tau_\theta(\cdot)$ \hfill \Comment{Neural network model}
            \item $S^*$ \hfill \Comment{Ground truth speed}
        \end{itemize}

        \For {$i=1, \dots$} \Comment{Unknown Environment Exploration}
            \State Obtain odom and sensor data from a given viewpoint
            \State Transform data to world coordinates
            \State Sample $N$ rays, each with $m$ stratified points
            \State Gather $n$ samples $\boldsymbol{\mathrm{q}}_{\{n\}} \subset \mathcal{Q}$ from $N$ rays (\ref{e-1})
            \State Compute expert speed $S^*_{\{n\}}$ for sampled points (\ref{e-1})
            \For {$j=1, \dots$} \Comment{ Online Training Epochs \ref{e-2}}
                \State Sample $\mathcal{B}_j \text{ from } \mathcal{M}$ \Comment{Batch train data}
                \State $\mathcal{B}_j=\mathcal{B}_j \cup (\boldsymbol{\mathrm{q}}_{\{n\}},S^*_{\{n\}})$ \Comment{Add new data}
                \State $\forall \big(q_s,q_g, S^*(q_s), S^*(q_g)\big) \in \mathcal{B}_j :$ 
                \State $\;\;\;\tau_\theta(q_s, q_g)$ \Comment{Predict factorized time Eq.~\ref{tf}}
                \State $\;\;\;S(q_s), S(q_g)$ \Comment{Predict speed by Eq.~\ref{predictspeed}}
                \State $\;\;\;l_j = L(S^*, S)$ \Comment{Compute loss by Eq.~\ref{trainloss}}
                \State $\;\;\;\theta = \theta - \nabla_{\theta} l_j$ \Comment{Update model parameters}
            \EndFor
            \State $\mathcal{M} = \mathcal{M} \cup (\boldsymbol{\mathrm{q}}_{\{n\}}, S^*_{\{n\}})$ \Comment{Update memory \ref{e-2}}
            \State Select Next Best Viewpoint from local time field \ref{e-3}
            \State Find a path to NBV by Eq.~\ref{plan}
        \EndFor
    \end{algorithmic}
\end{algorithm}

\begin{figure*}[!ht]
    \centering
        \centering
\includegraphics[width=0.98\textwidth,trim=0.0cm 4.4cm 0.0cm 4.4cm,clip]{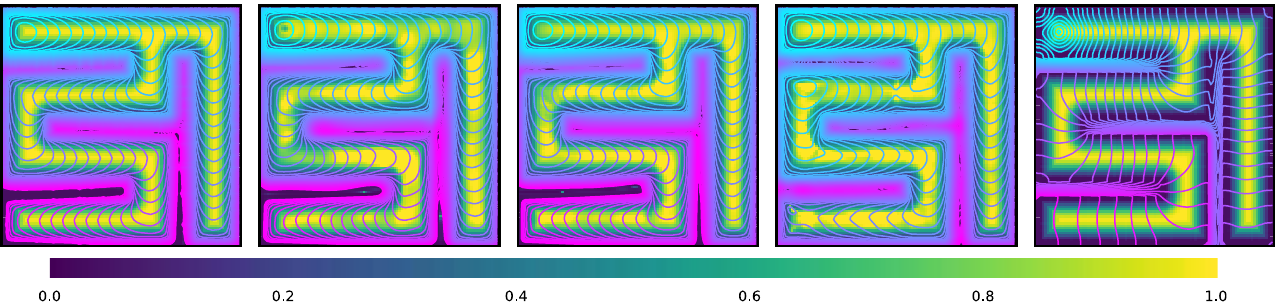}
    \put(-473,121){{ FMM}}
    \put(-368,121){{ Ours}}
    \put(-275,121){{ P-NTFields}}
    \put(-172,121){{ NTFields}}
    \put(-55,121){{ SD}}
    \caption{\justifying We compare our method with NTFields, P-NTFields, FMM, and { SD} on a maze environment for time field generation. Our method recovers the correct result within 15 seconds, whereas P-NTFields and NTFields take 18 and 10 minutes, respectively. { Although SD does not recover the exact time field, it provides a smooth global approximation, which inspired aspects of our architecture design.}}
    \label{fig:perfcompare}
\end{figure*}
\subsubsection{Next Viewpoint Selection}\label{e-3}

Upon training the neural network with the given local perception data, the next step is to select the subsequent viewpoint for acquiring new observations. Active exploration involves choosing the next optimal viewpoint to collect data using the robot's onboard sensor. As this work does not aim to develop novel exploration strategies, we utilize a standard approach to maximize scene coverage.

During exploration, we maintain an occupancy map. Regions of the map are marked as explored once the robot's sensor gathers sufficient information, while the remaining areas are labeled as unexplored.

To determine the next best view for exploration, we randomly sample collision-free points from the frontier of the explored occupancy map, considering these as candidates for the next best view. We then compute the arrival time to these points using our neural model from the robot's current location and select the point with the minimal arrival time. The path from the robot's current location to this next best view is calculated using the arrival time field, following the procedure described in Section IV-E.

It is important to note that our neural time field generator can be seamlessly integrated into any other exploration strategy. Since the next best viewpoints typically lie at the boundaries between explored and unexplored regions, the locally learned time fields can provide a path without the need for an external path planner.

\subsection{Active NTFields Online Learning Summary}
Algorithm \ref{alg:active_exploration} and Fig. \ref{fig:pipeline} provides our Active NTFields online training pipeline. The procedure begins with obtaining the observations from a given viewpoint and processing them to get sample points $\boldsymbol{\mathrm{q}}_{\{n\}}$ and their expert speed estimates $S^*_{\{n\}}$ (lines 3-7). These samples, along with some past samples from $\mathcal{M}$, are then used to train the network online for a few epochs (lines 9-15). Once trained, the next best viewpoint (NBV) is selected, and the path to NBV is generated using the learn arrival time field map (lines 17-18). The procedure is repeated until the unknown environment is explored and its arrival time field map is recovered. In our implementation, we choose iterator $j$ to go until 5 steps with the batch size of 2000 samples. Furthermore, we use AdamW \cite{Loshchilov2017DecoupledWD} with weight 1e-1 and learning rate 5e-4 as optimizer.

\section{Experiments}
\noindent This section presents our experiment analysis. We begin by comparing the training efficiency of our proposed method against prior physics-informed methods, i.e., NTFields \citep{ni2023ntfields} and P-NTFields \citep{Ni-RSS-23}, to demonstrate the effectiveness of our proposed approach in recovering arrival time fields online. Next, we quantitatively compare the inferred maps from our proposed approach against state-of-the-art mapping methods. Lastly, we demonstrate the effectiveness of learned arrival time field maps in performing fast motion planning. We also demonstrate our approach in real-world settings for performing both mapping and motion planning in a higher-dimensional C-space. All experiments were performed using a system with 3090 RTX GPU, Core i7 CPU, and 128GB RAM.

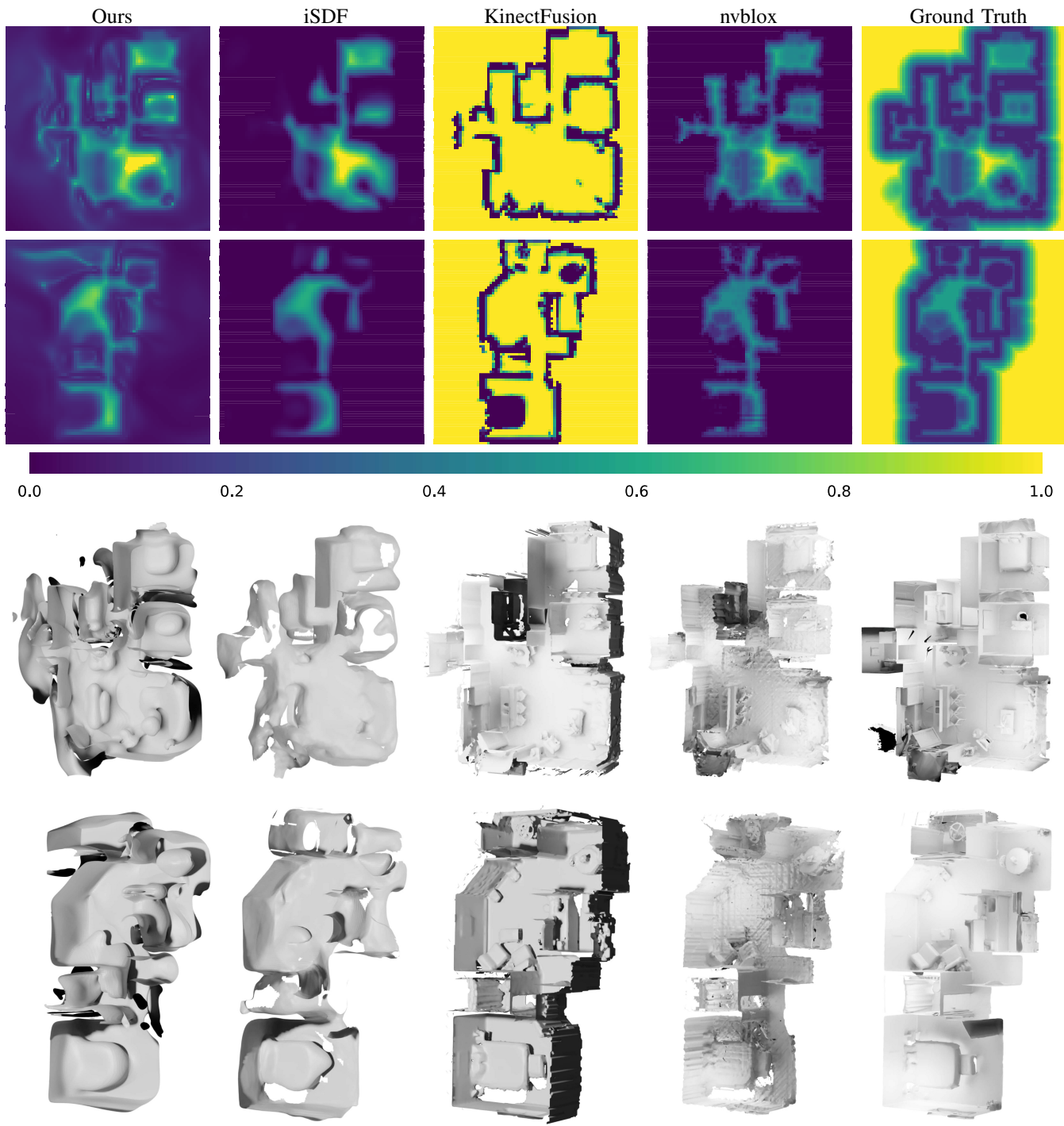
\begin{figure*}[!t]

\centering
        \centering
\begin{subfigure}[b]{2\linewidth}
        \includegraphics[width=0.49\textwidth,trim=0.0cm 2.2cm 0.0cm 2.2cm,clip]{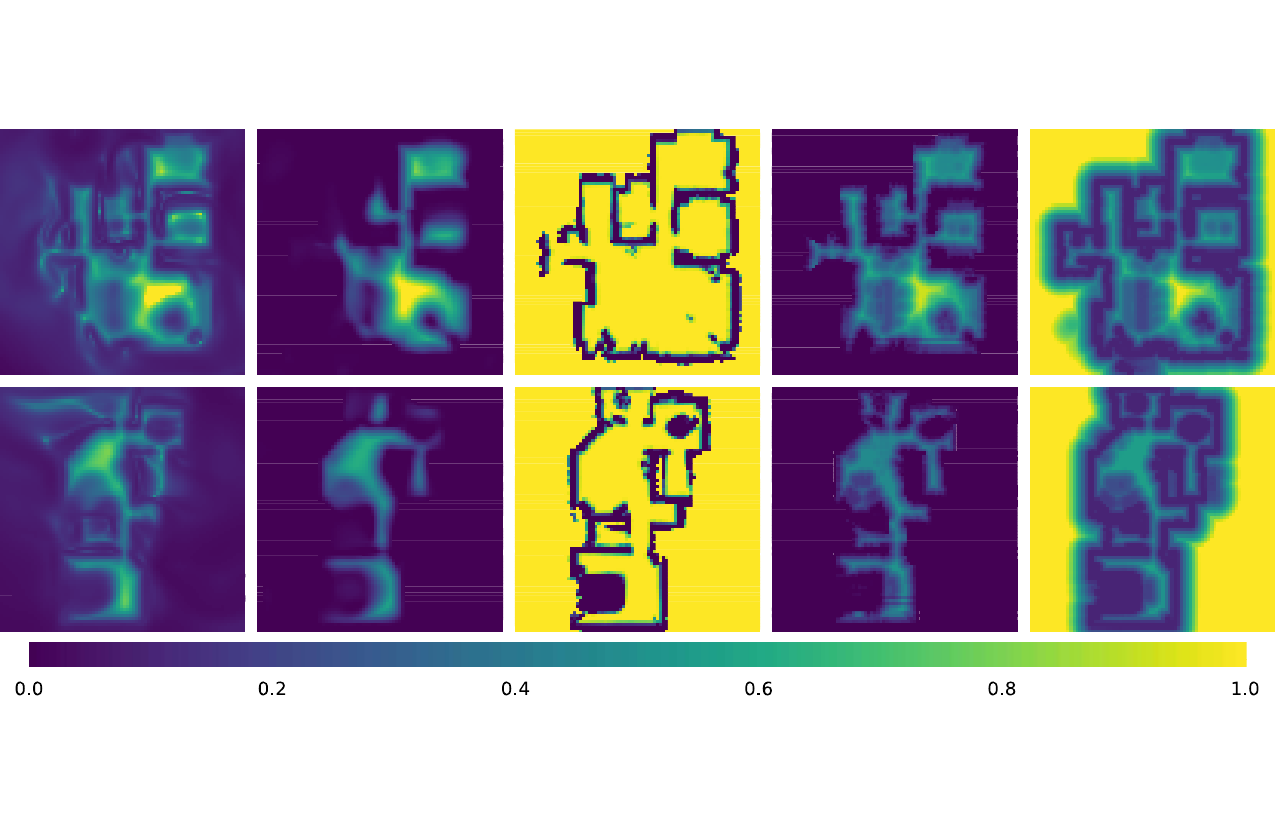}
        \put(-468,226){{ Ours}}
        \put(-367,226){{ iSDF}}
        \put(-281,226){{ KinectFusion}}
        \put(-172,226){{ { nvblox}}}
        \put(-78,226){{ Ground Truth}}
        \vspace{0.05in}
\end{subfigure}
\begin{subfigure}[b]{2\linewidth}
        \includegraphics[width=0.49\textwidth]{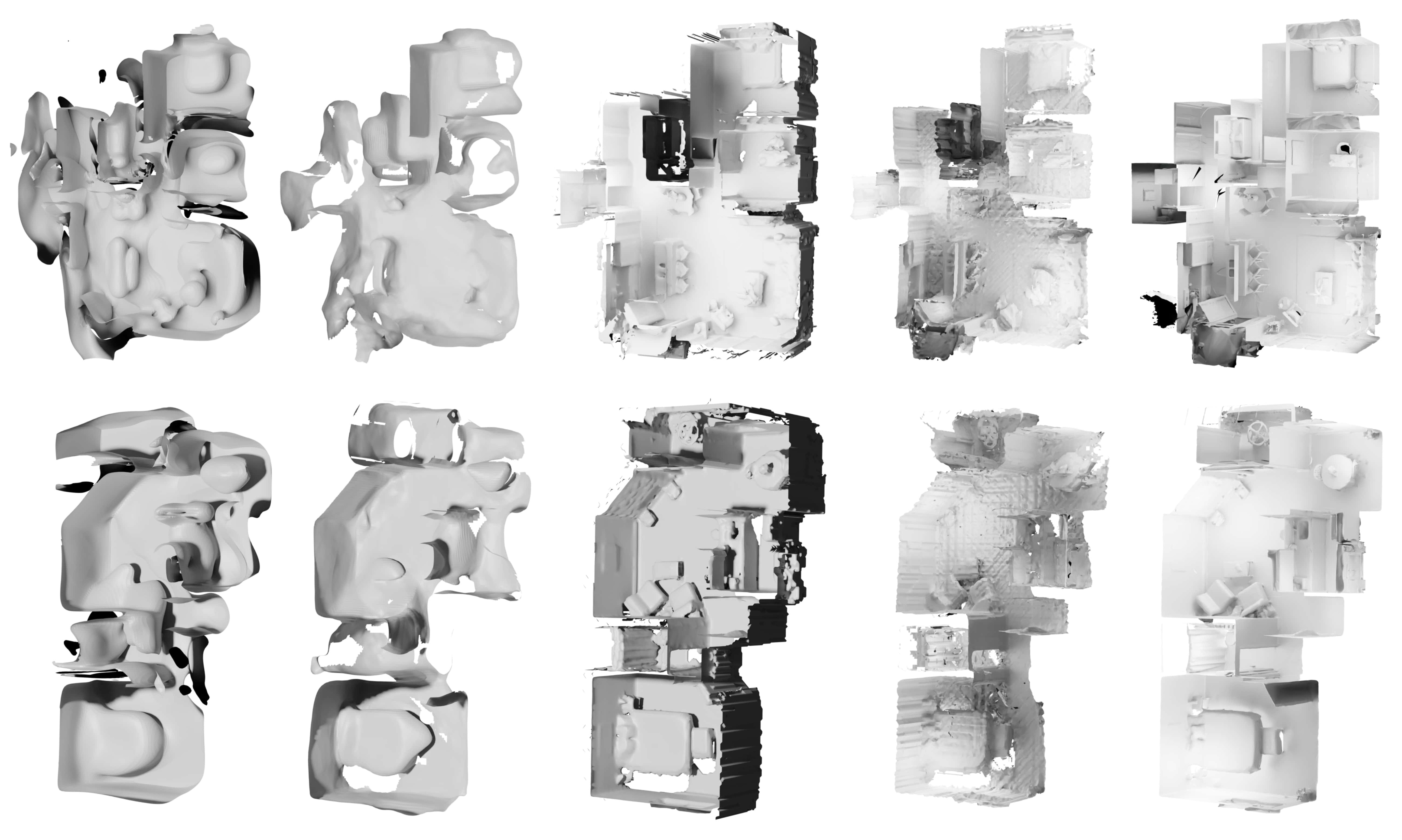}
        
\end{subfigure}
\caption{\justifying Comparison for mapping in Gibson environments. The figures show two indoor environments' SDF maps and zero-level-set mesh generated by our method, iSDF \citep{Ortiz-RSS-22}, KinectFusion \citep{newcombe2011kinectfusion}, and nvblox \citep{Millane2023nvbloxGI}.} 
\label{fig:mapping}
\vspace{-0.1in}
\end{figure*}

\subsection{Training Efficiency Analysis}

This section demonstrates our method performance over NTFields \citep{ni2023ntfields} and P-NTFields \citep{Ni-RSS-23}. Fig. \ref{fig:perfcompare} shows the speed and time fields of our method, NTFields, P-NTFields, { and SD} in comparison to FMM as ground truth. The color shows the speed fields and the contours show the arrival time fields from a start point. From Fig. \ref{fig:perfcompare} contours, our approach and P-NTFields get a similar result as FMM, while NTFields generate incorrect time fields. { Although the SD contours are not perfectly aligned with those of FMM, they offer a smooth and globally consistent approximation of the arrival time fields, which inspired elements of our feature design for learning the shortest arrival times.}

To quantitatively compare the results of all PINN methods, we sample points from their predicted arrival time and compute absolute error to the ground truth time field value given by the FMM. Our time fields have an error of $0.044 \pm 0.050$, P-NTFields have an error of $0.048 \pm 0.047$, and NTFields have an error of $0.44 \pm 0.42$. Note that the NTField error is almost ten times higher. Regarding training times, our method takes 15 seconds, while P-NTFields and NTFields take 18 minutes and 10 minutes, respectively. While our method and P-NTFields have similar results, the latter is about 72 times faster than the former in terms of training speed. Hence, this performance comparison validates our method's effectiveness and its suitability for online continual learning in mapping tasks.

\subsection{Mapping Comparison}

This section demonstrates our method performance on mapping tasks over KinectFusion \citep{newcombe2011kinectfusion}, iSDF \citep{Ortiz-RSS-22}, and nvblox \citep{Millane2023nvbloxGI}. We use eight Gibson environments \citep{pmlr-v164-li22b} to demonstrate the 3D mapping of indoor scenes. Due to the three baselines requiring a given trajectory, we first run our active exploration strategy to obtain a series of depth images and camera poses. With these raw data, all methods reconstruct the maps of all indoor scenes. 
We compared all {other} methods over SDF quality and mapping efficiency, as our speed fields can be seen as truncated SDF. 
In Fig. \ref{fig:mapping}, we present two environments of SDF and their zero-level-set mesh. Our method, iSDF, { and nvblox} capture a similar result as the ground truth, while KinectFusion can only reconstruct the SDF very close to the obstacles as it cannot support large truncated regions. However, from the zero-level-set mesh, KinectFusion  {  and nvblox} can reconstruct high-quality obstacles, whereas our method and iSDF cannot capture many details. Note that we are focusing on the downstream motion planning applications, and rough obstacle details are fine for collision avoidance, as also validated by our motion planning experiments in the next section. Table~\ref{table:mapping} shows the SDF error and reconstruction time. To compute the SDF error, we sample points within the truncated region and calculate the absolute error of SDF value against ground truth results. 
In the table, it can be seen that our SDF error is similar to iSDF. Regarding mapping efficiency, although our method is slower than the baseline methods as it gathers informative features, our mapping time is comparable with baseline methods and suitable for online tasks.

\begin{table}[h]
\centering
\scalebox{1}{ 
\begin{tabular}{cccc}
\toprule
\multirow{2}{*}{SDF} & \multicolumn{3}{c}{Performance Metrics} \\ 
\cmidrule{2-4}
& \multicolumn{1}{c}{SDF Error $\downarrow$} & \multicolumn{1}{c}{Frame Time (s) $\downarrow$} & \multicolumn{1}{c}{Mapping Time (s) $\downarrow$} \\ 
\midrule
Ours & $0.09 \pm 0.10$ & $2.12 \pm 0.06$ & $149.74 \pm 18.24$ \\
iSDF & $0.09 \pm 0.11$ & $0.94 \pm 0.02$ & $69.90 \pm 7.57$ \\
KFusion & $0.47 \pm 0.27$ & $1.71 \pm 0.01$ & $126.08 \pm 16.06$ \\
nvblox & $0.07 \pm 0.07$ & $0.12 \pm 0.00$ & $9.16 \pm 3.70$ \\
\bottomrule
\end{tabular}}
\caption{Quantitative comparison of our method against iSDF \citep{Ortiz-RSS-22}, Kinect Fusion (KFusion) \citep{newcombe2011kinectfusion}, and nvblox \citep{Millane2023nvbloxGI} in mapping the Gibson environments.}
\label{table:mapping}
\end{table}

\begin{table}[!t]
\centering
\scalebox{1}{ 
\begin{tabular}{cccc}
\toprule
\multirow{2}{*}{Methods} & \multicolumn{3}{c}{Performance Metrics} \\ 
\cmidrule{2-4}
& \multicolumn{1}{c}{Time (sec) $\downarrow$} & \multicolumn{1}{c}{Length $\downarrow$} & \multicolumn{1}{c}{SR (\%) $\uparrow$} \\ 
\midrule
Ours (G) & $0.01 \pm 0.05$ & $0.48 \pm 0.43$ & 98.28 \\
LazyPRM (C) & $0.34 \pm 1.54$ & $0.41 \pm 0.22$ & 99.28 \\
iSDF+MPOT (G) & $1.42 \pm 0.52$ & $0.62 \pm 0.27$ & 91.43 \\
KFusion+MPOT (G) & $0.28 \pm 0.08$ & $0.55 \pm 0.16$ & 93.71 \\
nvblox+MPOT (G) & $0.27 \pm 0.05$ & $0.58 \pm 0.14$ & 96.57 \\
iSDF+RRTConnect (G) & $1.44 \pm 2.13$ & $0.35 \pm 0.20$ & 83.85 \\
KFusion+RRTConnect (G) & $0.89 \pm 1.36$ & $0.38 \pm 0.21$ & 91.57 \\
nvblox+RRTConnect (G) & $0.51 \pm 0.99$ & $0.39 \pm 0.21$ & 88.71 \\
KFusion+RRTConnect (C) & $0.20 \pm 0.48$ & $0.40 \pm 0.22$ & 94.71 \\
nvblox+RRTConnect (C) & $0.27 \pm 0.46$ & $0.40 \pm 0.22$ & 96.29 \\
\bottomrule
\end{tabular}}
\caption{Comparison for motion planning in eight Gibson environments. For each environment, 100 randomly sampled starts and goals near obstacles are selected for evaluation. We load grid-based maps in both CPU (C) and GPU (G) for RRTConnect-based approaches.} 
\label{table:mp_gib}
\end{table}
\begin{figure}[!ht]
\centering
        \centering
\begin{subfigure}[b]{4\linewidth}
        \includegraphics[width=0.24\textwidth, trim=14.3cm 10.3cm 17.2cm 4.2cm,clip]{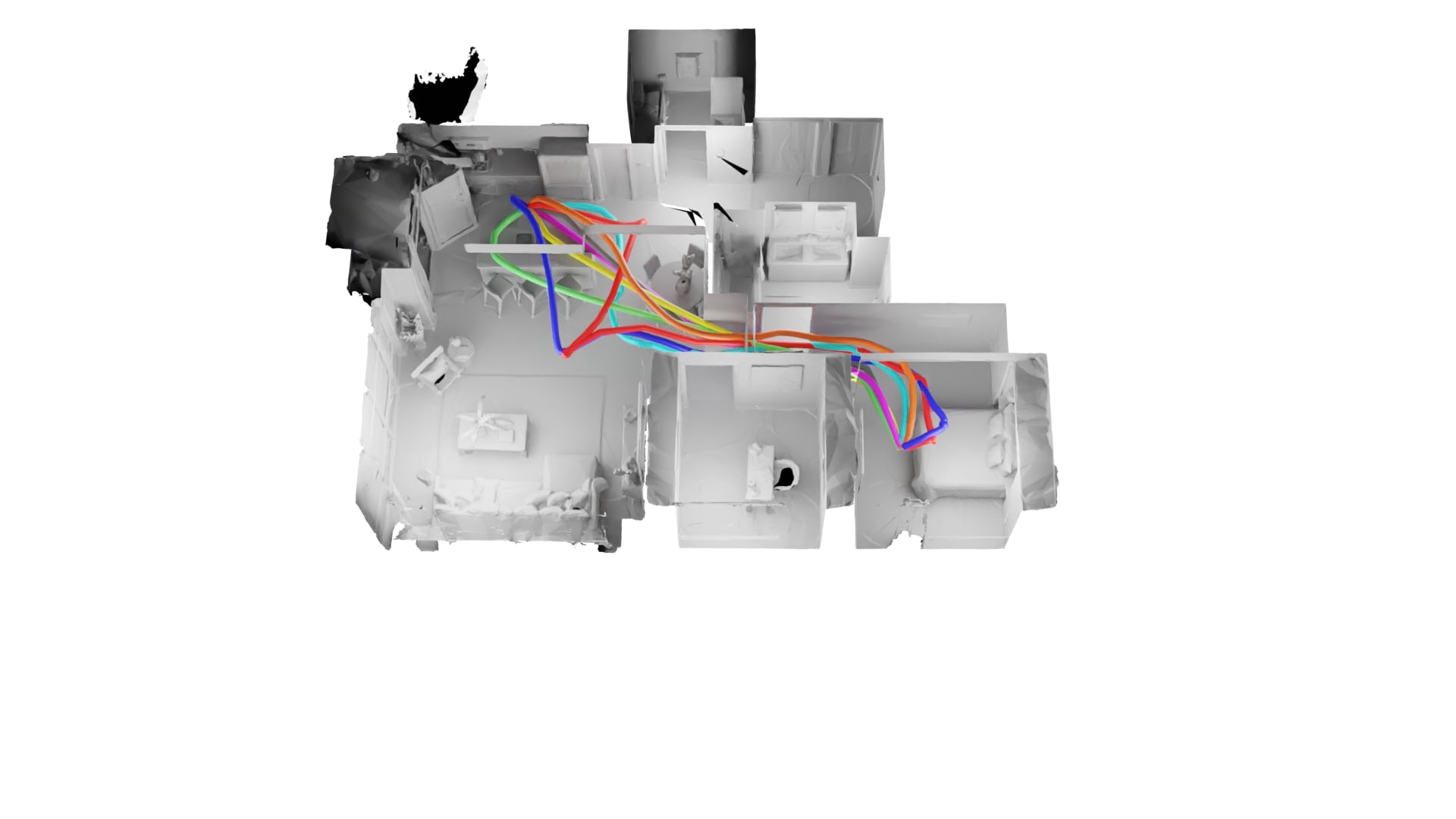}
\end{subfigure}
\begin{subfigure}[b]{4\linewidth}
        \includegraphics[width=0.24\textwidth,trim=14.3cm 4.7cm 2.2cm 0.2cm,clip]{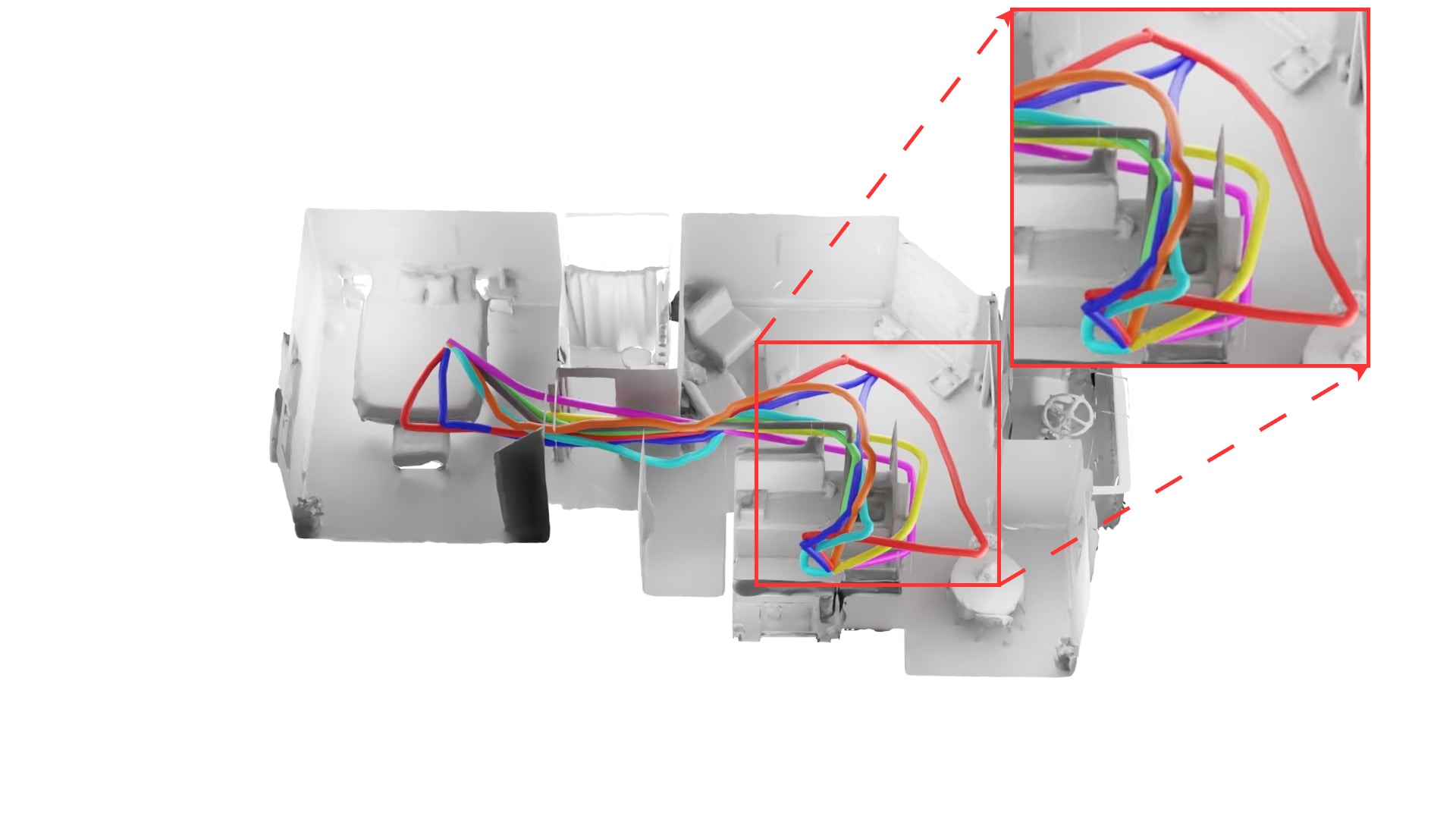}       
\end{subfigure}
\caption{\justifying Comparison for motion planning in Gibson environments.
The figures show five paths generated by our method (orange), {LazyPRM (gray),} iSDF+MPOT (red), KinectFusion+MPOT (cyan), iSDF+RRTConnect (green), KinectFusion+RRTConnect (yellow), {nvblox+MPOT (blue), and nvblox+RRTConnect (pink)}. The table shows statistical results on 8$\times$100 different starts and goals in eight Gibson environments. {We load grid-based maps in both CPU and GPU in RRTConnect to test the performance. {The highlighted region illustrates that MPOT and RRTConnect may discover longer paths belonging to a different homotopy class.}} }
\label{fig:planning}
\vspace{-0.25in}
\end{figure}
\begin{figure*}[!ht]
    \centering
    \includegraphics[width=0.8\textwidth]{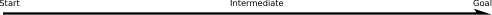}
    \\[0.1cm]
    \includegraphics[width=0.01\textwidth,trim=0.0cm -0.75cm 0.0cm 0.0cm,clip]{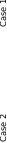}
    \begin{subfigure}[b]{0.97\textwidth}
        \centering
        \includegraphics[width=0.19\textwidth]{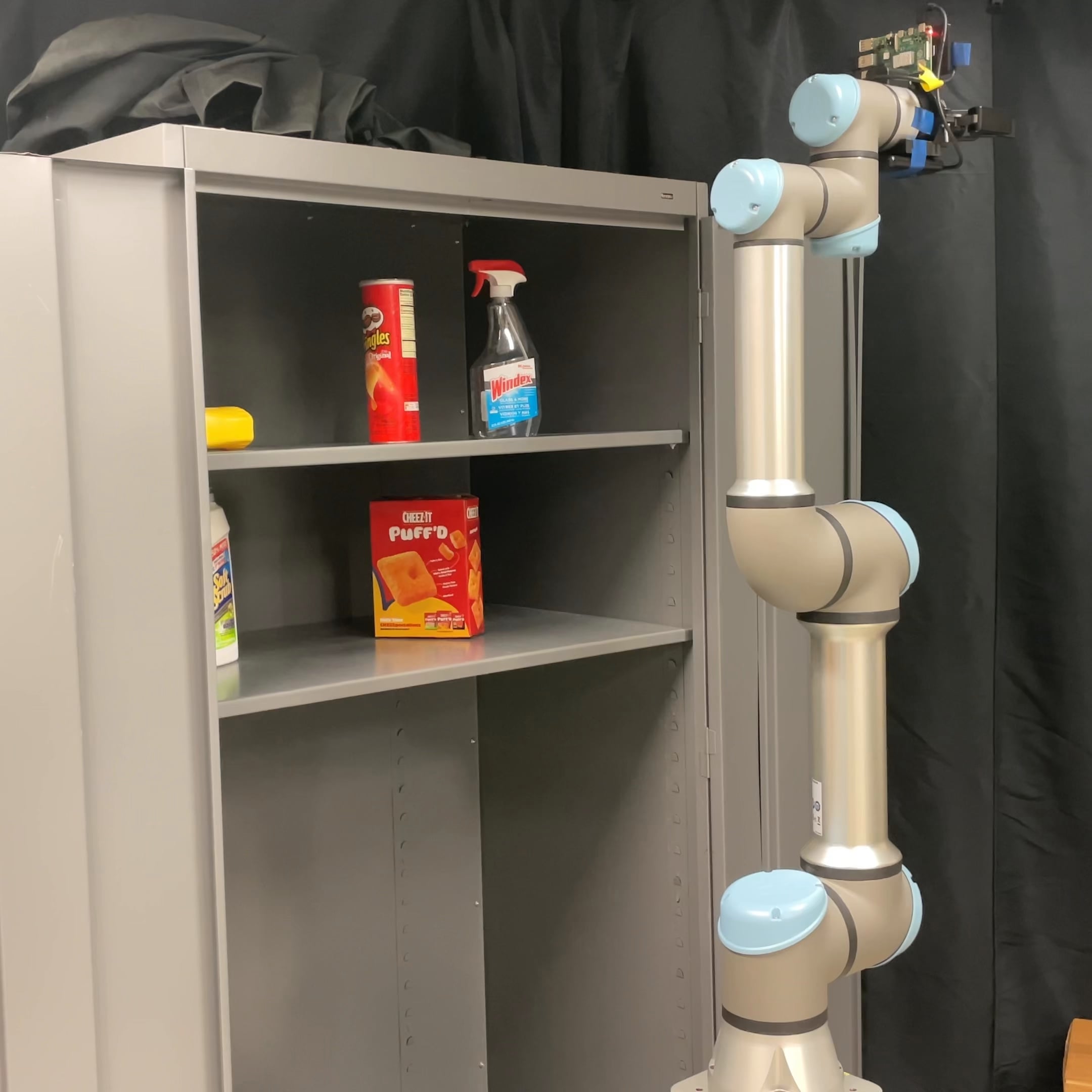}
        \includegraphics[width=0.19\textwidth]{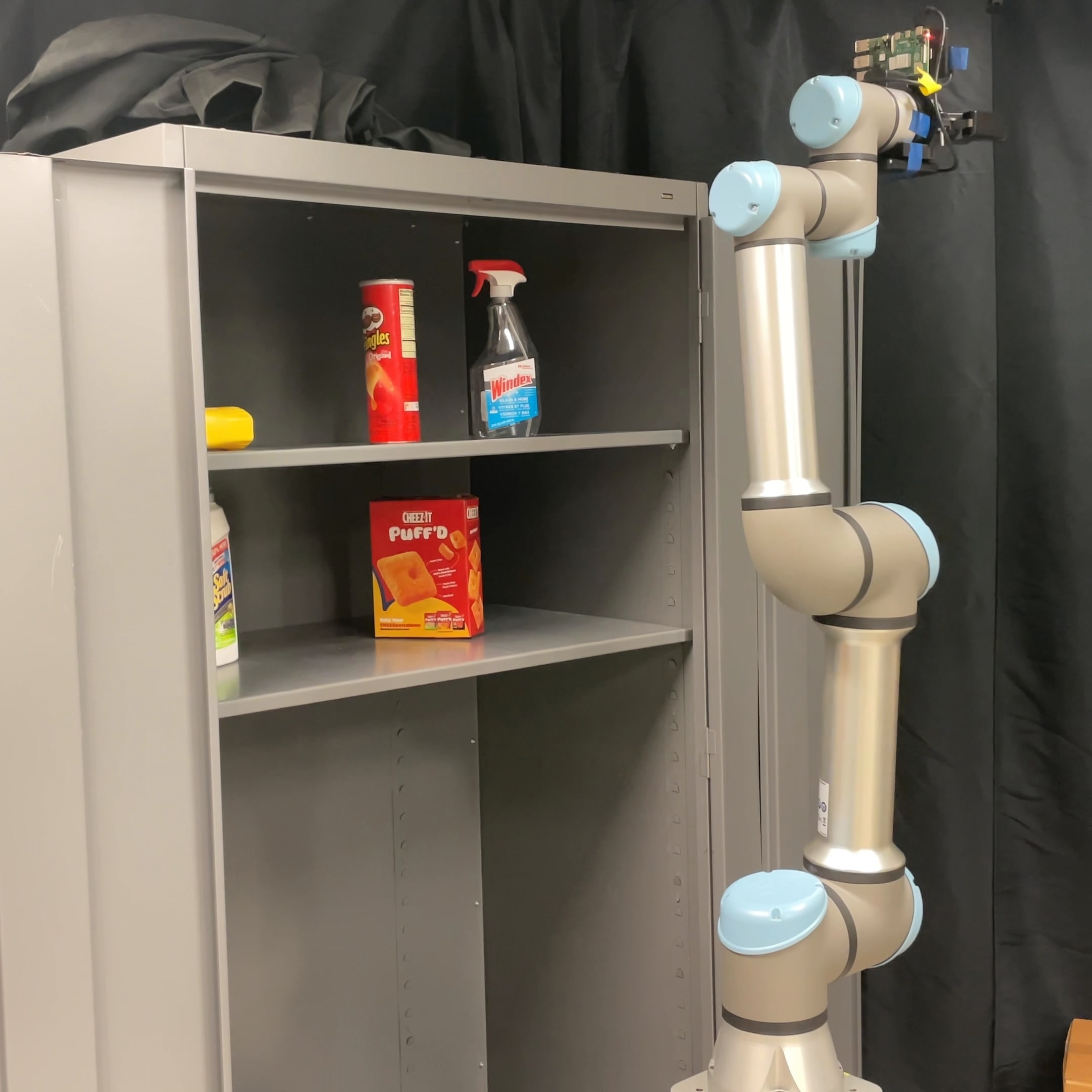}
        \includegraphics[width=0.19\textwidth]{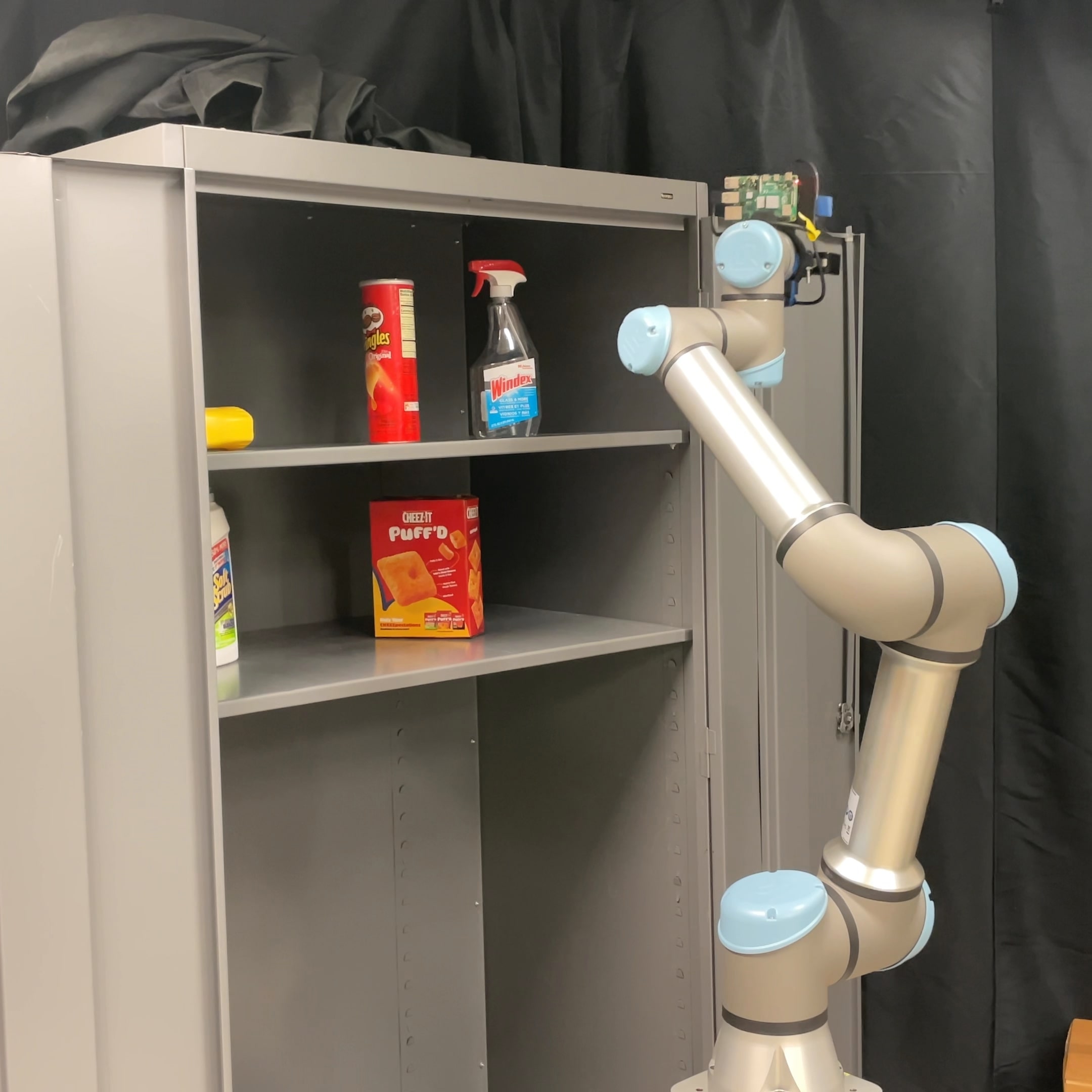}
        \includegraphics[width=0.19\textwidth]{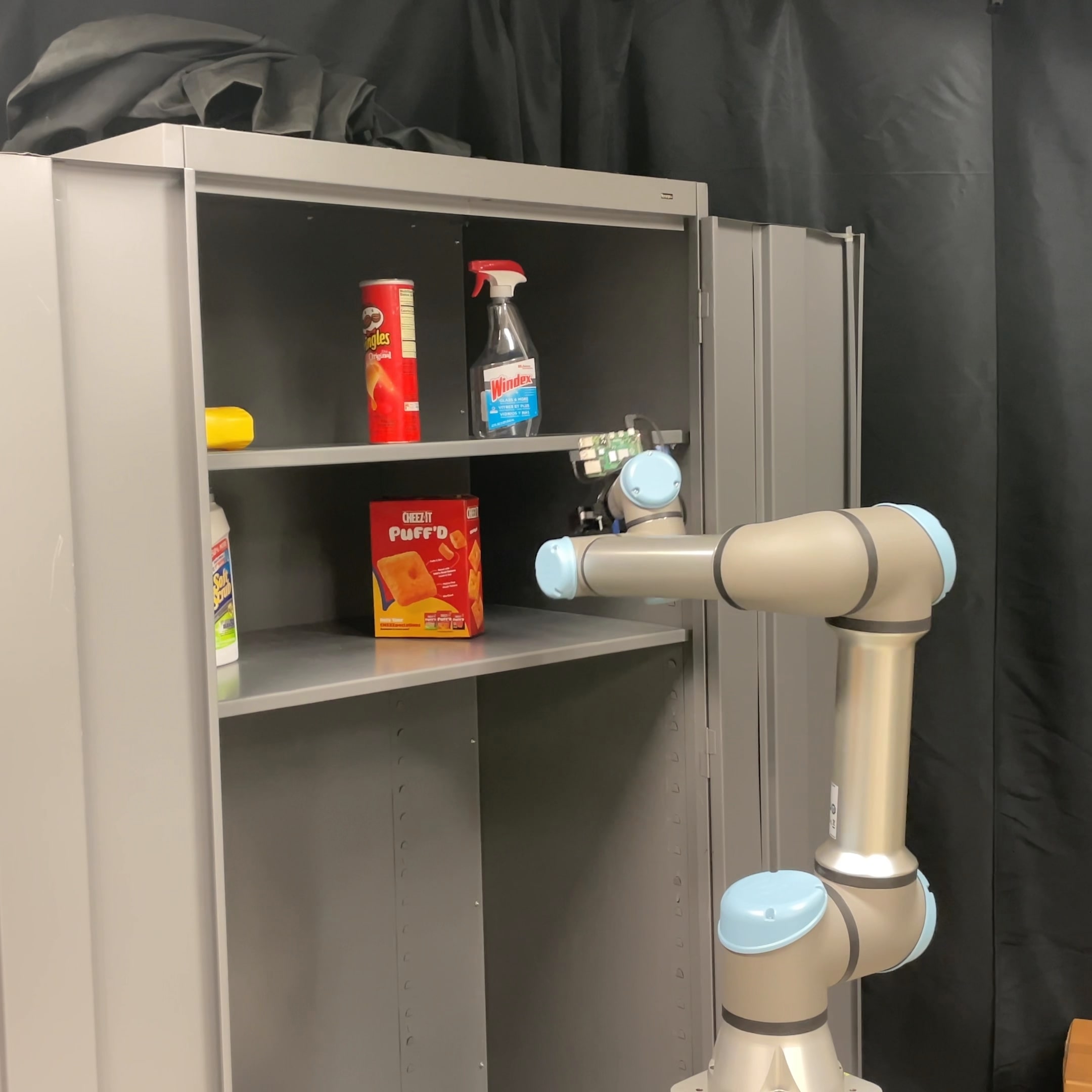}
        \includegraphics[width=0.19\textwidth]{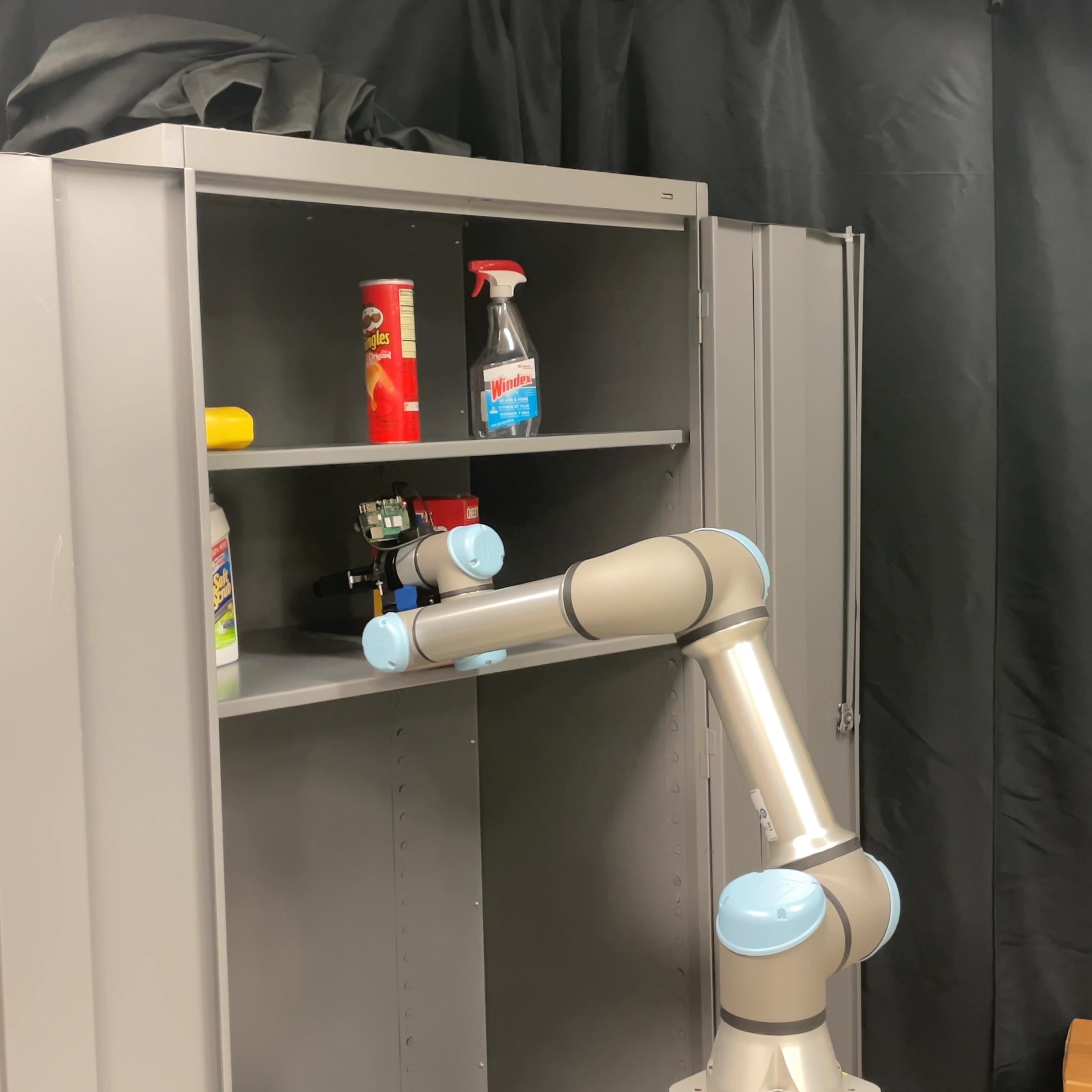}
        \\[0.1cm]
        \includegraphics[width=0.19\textwidth,trim=0.0cm 6cm 0.0cm 6cm,clip]{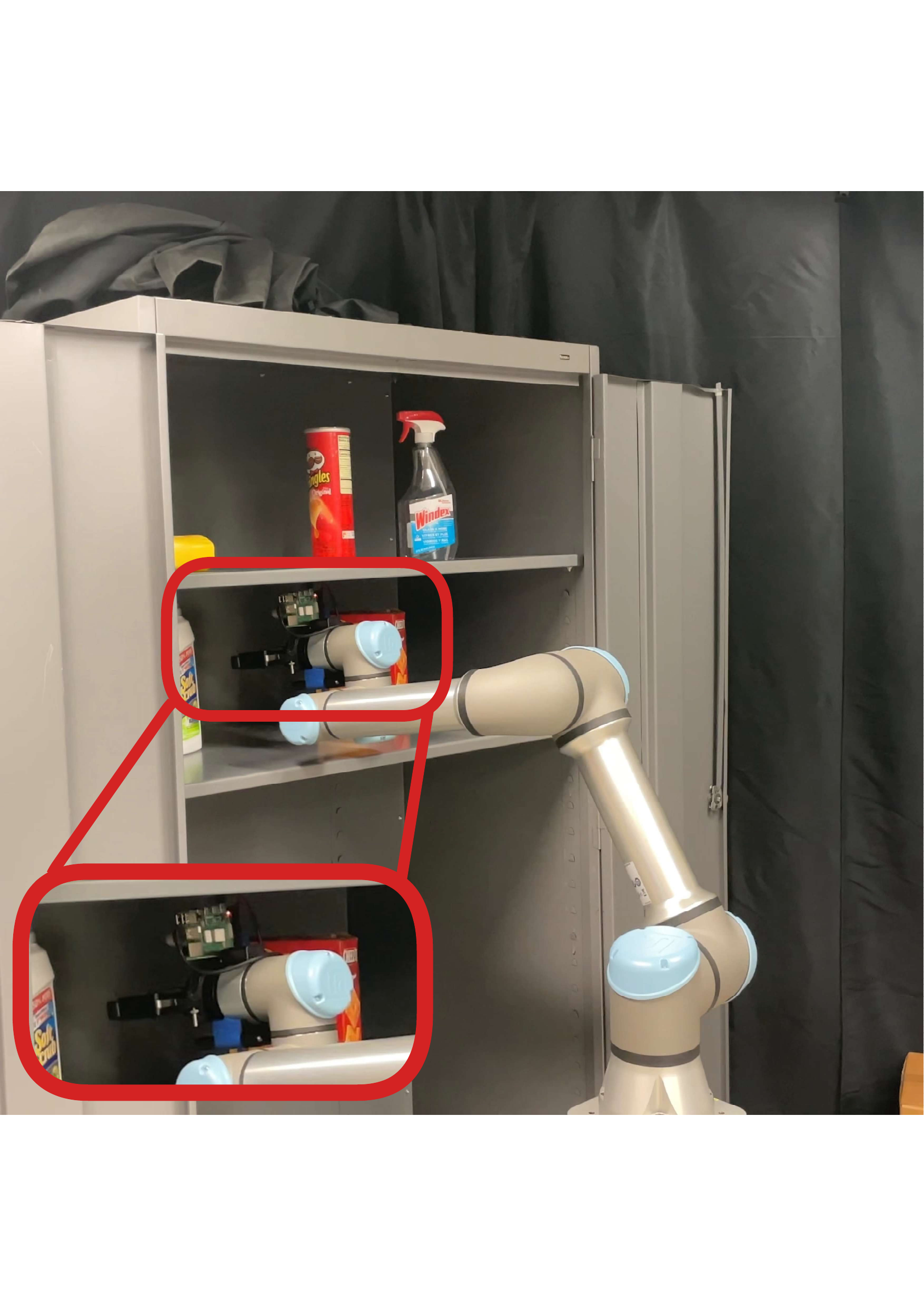}
        \includegraphics[width=0.19\textwidth]{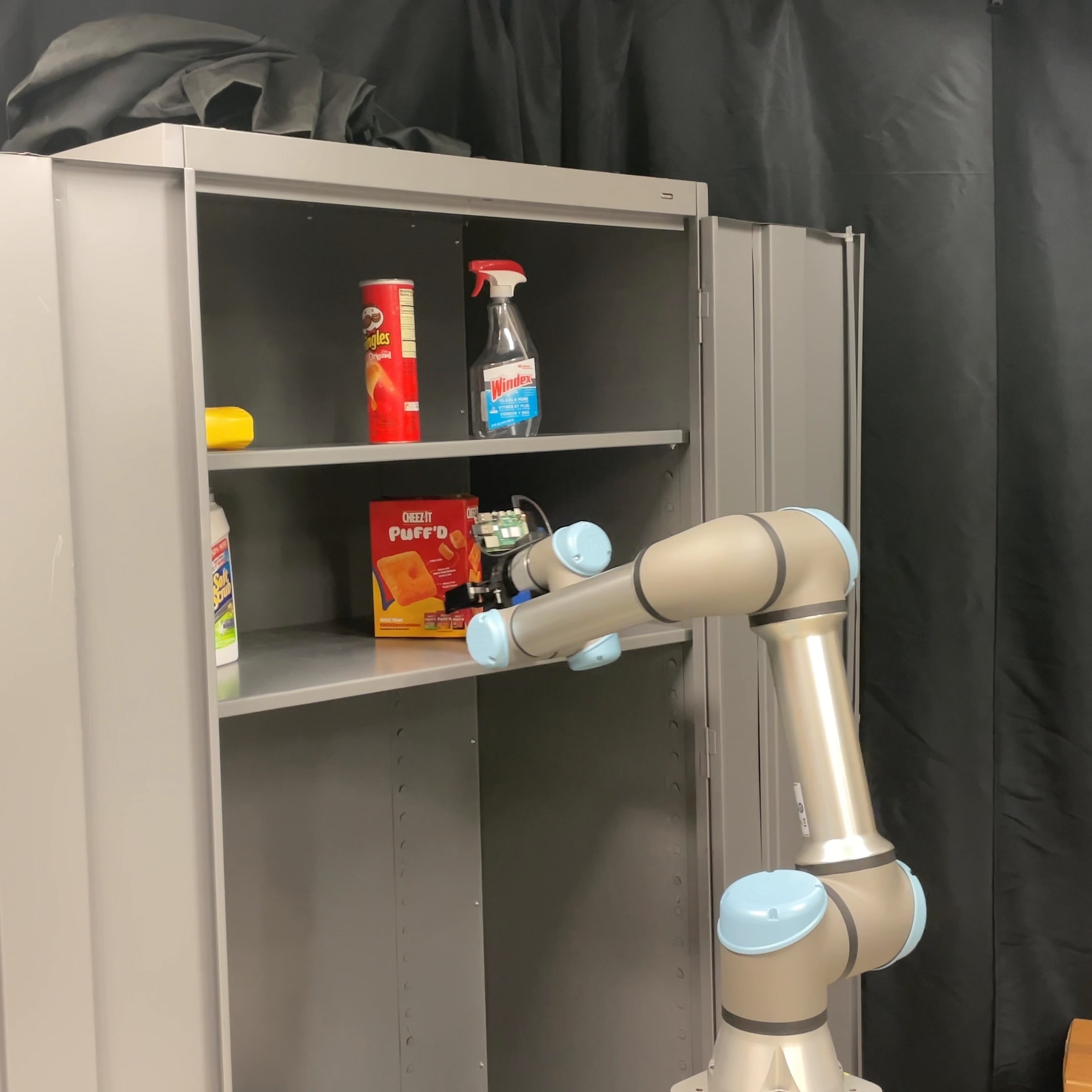}
        \includegraphics[width=0.19\textwidth]{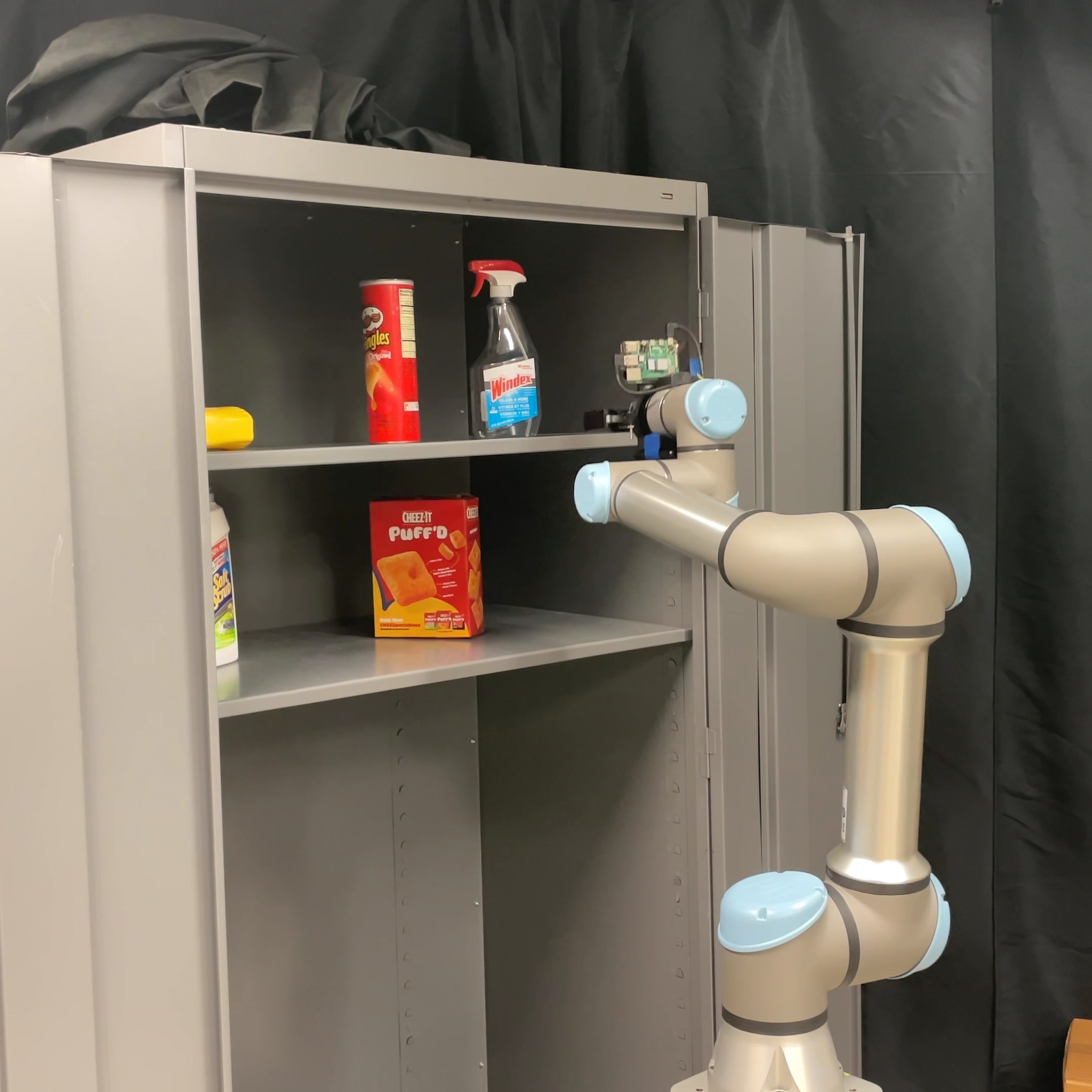}
        \includegraphics[width=0.19\textwidth]{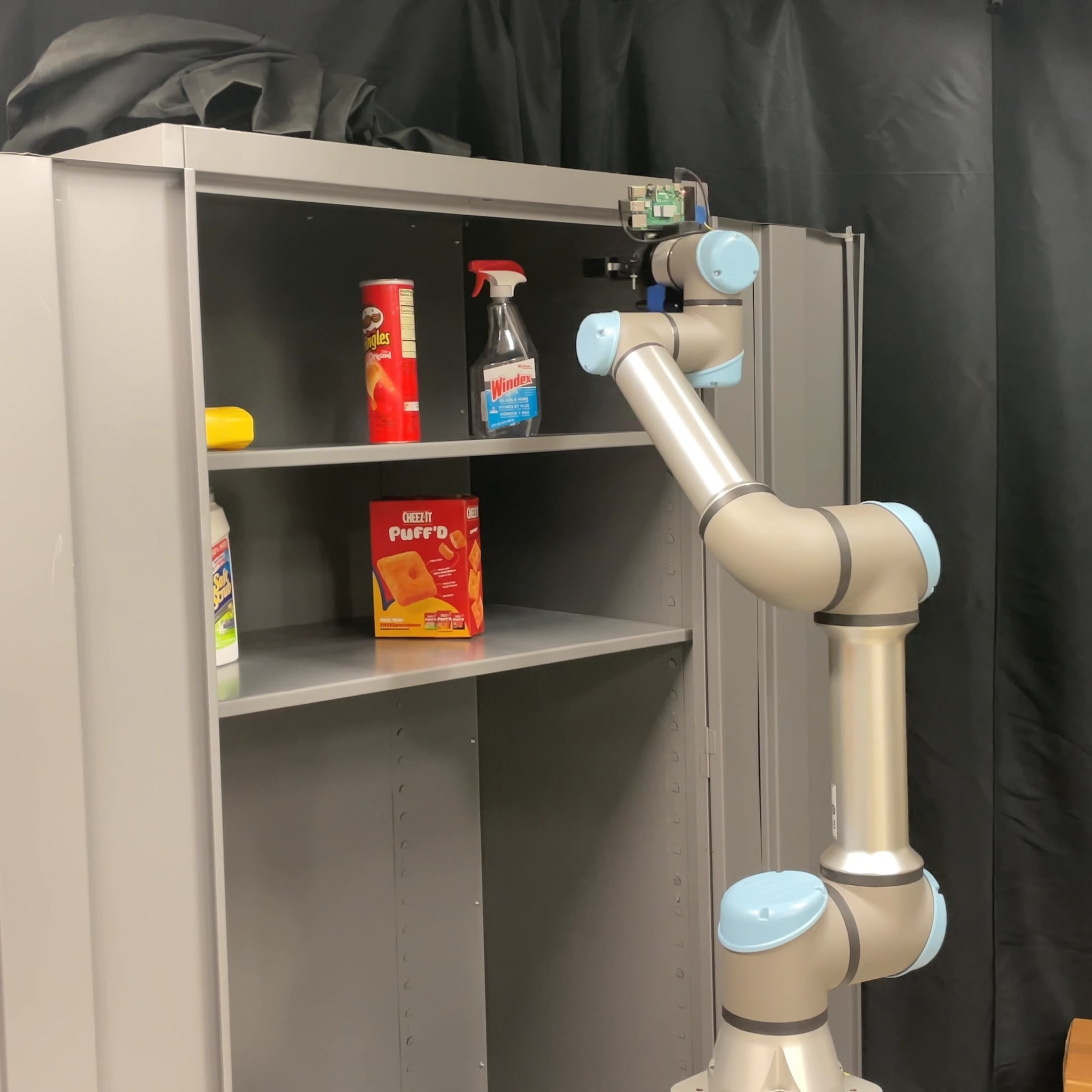}
        \includegraphics[width=0.19\textwidth]{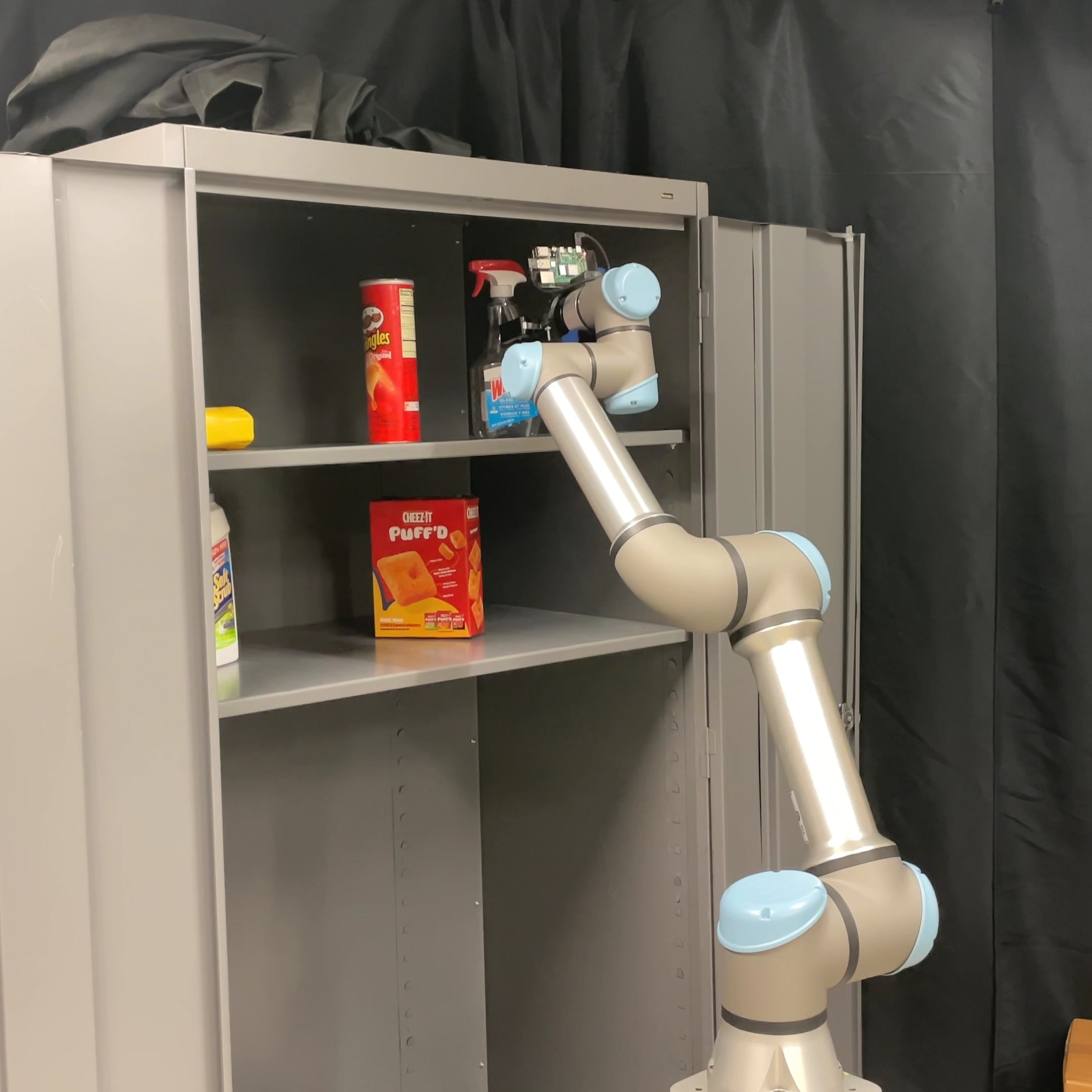}
    \end{subfigure}

    \caption{\justifying Real-world cabinet environment: Our method reconstructed the arrival time fields using the in-hand camera. The top and bottom rows show two cases of our motion planning problems in this confined environment. The first case shows the manipulator going into the cabinet, and the second case shows the manipulator crossing the cabinet's middle level to reach the given target at the top level. {We highlight the end effector to show the narrow passage condition of Case 2.} In this scenario, our method takes only 0.02 seconds to find the path, {whereas LazyPRM (CPU) takes 3.72 seconds, KinectFusion+RRTConnect (CPU) takes 4.15 seconds, and nvblox+MPOT (GPU) takes 4.63 seconds.}}\vspace{-0.2in}
    \label{fig:arm_real}
\end{figure*}
\subsection{Motion Planning Comparison}
This section demonstrates our method performance on motion planning tasks over MPOT \citep{le2023accelerating}, RRTConnect with smoothing \citep{kuffner2000rrt}, and online LazyPRM with smoothing\citep{bohlin2000path}.

The MPOT is the most recent and best available method that outperformed various state-of-the-art motion planning methods. Furthermore, we use the reconstructed maps of eight Gibson environments \citep{pmlr-v164-li22b} from our mapping results (Section V-B). { Hence, two planners, MPOT and RRTConnect, use the reconstructed map from iSDF, KinectFusion, and nvblox. {MPOT is a GPU-based method, and we only load the grid map to GPU. Additionally, MPOT was run in batches of 100 trajectories which we select to balance the trade-off between its success rate and computation time.} RRTConnect is a CPU-based method, and we load the grid map on both GPU and CPU. LazyPRM uses its own reconstructed roadmap and uses graph search with smoothing to find the path.} In contrast, our method can directly generate motion planning paths by following Eq. \ref{plan}, demonstrating the effectiveness of arrival time field mapping. 

We randomly choose 100 start and goal points in each environment and run the above-mentioned motion planning methods to get the final paths. We compare the path length, computational time, and success rate. {The collision checking is performed using the ground-truth map to validate the generated paths and compute the success rate. We also set a time limit of 10 seconds for sampling-based methods LazyPRM and RRTConnect, after which the case is considered a failure.} 

Fig. \ref{fig:planning} shows the paths where our method and KinectFusion+MPOT generate short and smooth paths, but iSDF+MPOT { and nvblox+MPOT} generate damping results. RRTConnect only considers the SDF zero-level-set; thus, all RRTConnect generate similar results. Table~\ref{table:mp_gib} presents the statistical results. Our method's computational time is about 30 times faster than MPOT and achieves 98\% success rate. In addition, our method generates smooth paths with a safe margin to obstacles, while RRTConnect generates paths close to the obstacle, resulting in relatively shorter lengths. 

{ All sampling-based methods failed in some cases primarily due to exceeding the time limit. These methods rely on sampling strategies that can require significant time to explore complex environments and find a global collision-free path, especially when narrow passages are present. In contrast, MPOT failed primarily because it was unable to find a global collision-free path, even when sufficient time was available, as it depends on local optimization, which may get trapped in local minima without a global guarantee.

Our method, like MPOT, also failed in some cases because it could not find a global collision-free path, but it demonstrated higher robustness and faster convergence. To clarify the success rate further, failures for sampling-based methods occurred due to exceeding the time limit, while for MPOT and our method, failures were due to the inability to find a feasible path without collisions.
}

\begin{table}[t]
\centering
\scalebox{1}{
\begin{tabular}{cccc}
\toprule
\multirow{2}{*}{Methods} & \multicolumn{3}{c}{Performance Metrics} \\ 
\cmidrule{2-4}
& \multicolumn{1}{c}{Time (sec) $\downarrow$} & \multicolumn{1}{c}{Length $\downarrow$} & \multicolumn{1}{c}{SR (\%) $\uparrow$} \\ 
\midrule
Ours (G) & $0.03 \pm 0.01$ & $2.25 \pm 0.72$ & 91.00 \\
LazyPRM (C) & $1.36 \pm 0.81$ & $3.05 \pm 0.89$ & 87.00 \\
iSDF+MPOT (G) & $3.63 \pm 1.29$ & $4.04 \pm 0.77$ & 89.00 \\
KFusion+MPOT (G) & $2.18 \pm 0.06$ & $3.99 \pm 0.48$ & 79.00 \\
nvblox+MPOT (G) & $2.11 \pm 0.39$ & $4.15 \pm 0.56$ & 88.00 \\
iSDF+RRTConnect (G) & $3.92 \pm 8.20$ & $2.33 \pm 1.04$ & 89.00 \\
KFusion+RRTConnect (G) & $2.71 \pm 4.56$ & $2.18 \pm 0.83$ & 85.00 \\
nvblox+RRTConnect (G) & $3.16 \pm 3.53$ & $2.30 \pm 1.09$ & 84.00 \\
KFusion+RRTConnect (C) & $1.64 \pm 1.35$ & $2.48 \pm 1.21$ & 82.00 \\
nvblox+RRTConnect (C) & $1.90 \pm 5.63$ & $2.32 \pm 0.96$ & 82.00 \\
\bottomrule
\end{tabular}}
\caption{Comparison for motion planning in the UR5e Manipulator in cabinet environment with 100 randomly sampled start and goals near obstacles. We load grid-based maps in both CPU (C) and GPU (G) in RRTConnect to test the performance.}
\label{table:arm}\vspace{-0.25in}
\end{table}\vspace{-0.1in}

\subsection{Real World Experiments}
This section presents our real-world experiments with differential drive TurtleBot4 and 6-DOF UR5E robots.
\subsubsection{TurtleBot4 in indoor environment}

In this experiment, a TurtleBot4 robot equipped with a RealSense2 camera is used to explore an indoor environment with furniture. The camera poses are from the TurtleBot4 odometry, while depth images are from the RealSense2 camera. Fig. \ref{fig:exploration} shows our method of incrementally constructing the map with incoming frames. Furthermore, our method uses the reconstructed arrival time field in the partially observed environment to reach the next viewpoint for sensing. Thus, it does not require any external motion planner during mapping. In Fig. \ref{fig:exploration}, the color represents the speed fields, and the contour lines indicate the time fields. It takes 65 seconds for the robot to actively explore and reconstruct the arrival time field of the whole environment. These results highlight the capability of our approach in mapping a real-world indoor environment.

Once the map reconstruction is done, any start and goal pair in the map can be fed into the neural network, and a path is generated in a bidirectional way using Eq. \ref{plan}. To evaluate our method of motion planning performance, we randomly sampled 100 starts and goals in the real environment. On average, our method computation times remained around 0.02 seconds with a 98\% success rate. Finally, the bottom row of Fig. \ref{fig:exploration} depicts an example path. We pick the start location near the sofa and the goal location behind a chair. It takes only 0.02 seconds to generate a valid smooth trajectory, avoiding collision with all the furniture.

\subsubsection{UR5e Manipulator in cabinet environment}
{  We use the UR5e robotic arm with a hand-held RealSense2 camera to navigate a realistic cabinet setting. Our approach generates an arrival time field map in 6 DOF C-Space and completes this process in 115 seconds. By comparison, alternative workspace maps reconstruction methods like iSDF, KinectFusion, and nvblox require significantly less time -- 6 seconds, 2.47 seconds, and 0.55 seconds, respectively. Note that the time difference is primarily because our method maps the higher-dimensional (6D) C-Space, whereas baseline methods map the 3D workspace. The baseline SDF methods cannot scale to C-Space mapping. Furthermore, the LazyPRM constructed from incremental local observations takes 162 seconds to build the roadmap in C-Space and is slower than our method. 

For motion planning, we used the reconstructed map and conducted tests using 100 randomly selected start and goal pairings near obstacles. Our method demonstrated a quick average planning time of 0.03 seconds with a 91\% success rate, as detailed in Table~\ref{table:arm}, achieving a computational time approximately 40 times faster than that of LazyPRM. Fig. \ref{fig:arm_real} showcases two instances of motion planning within this environment. The first scenario illustrates the manipulator initiating movement from an open area and entering the cabinet, while the second shows the manipulator navigating across the cabinet’s levels to reach a specified target. Notably, the manipulator begins from a confined position deep within the cabinet in Case 2. 
Furthermore, in this particular scenario (case 2), our method was able to find a solution in just 0.02 seconds, whereas  LazyPRM (CPU) tool 3.72 seconds, KinectFusion+RRTConnect (CPU) takes 4.15 seconds, and nvblox+MPOT (GPU) takes 4.63 seconds. These experiments demonstrate the scalability of our approach to high-dimensional C-Space and real-world confined environments.
}

\section{Conclusions and Future Work}
This paper presents Active Neural Time Fields that actively explore and learn arrival time field mapping of unknown environments. The arrival time fields allow extremely fast motion planning without requiring any computationally expensive tools. We demonstrate the effectiveness of our approach on a differential drive robot in mapping and navigating eight Gibson and real-world kitchen environments. We also showcase our approach with a 6-DOF robot arm with {a hand-held} camera for mapping and solving motion planning in a real-world narrow passage, cabinet-like environment. We compare our framework against state-of-the-art mapping and motion planning approaches. The mapping results demonstrate that our approach can map the environments with arrival time field features that can directly provide robot navigation paths between any start and goal robot configurations. The motion planning comparison shows at least 40$\times$ speed enhancement over the best available method. 

{ In our future work, we plan to investigate continual learning approaches for training our neural network to explore large-scale environments. We also aim to explore ensemble-based neural approaches to address the spectral bias issue of neural networks that hinders them from tackling high-frequency features that also appear frequently in very large environments. A parallel development to our method is a new line of work related to
hardware-accelerated motion planning that also aims to expedite motion planning \cite{thomason2024motions}, \cite{Ramsey-RSS-24}. These methods leverage traditional occupancy maps. Therefore, in our future work, we also aim to explore integrating our mapping feature with strategies in hardware-accelerated motion planning to investigate if motion planning speed can be further improved beyond microseconds. Lastly, we also plan to extend our framework to outdoor mapping scenarios, which we believe can open new ways to solve autonomous driving tasks without needing expert demonstrations. }

\bibliographystyle{plainnat}
\bibliography{references}

\begin{thebibliography}{75}
\providecommand{\natexlab}[1]{#1}
\providecommand{\url}[1]{\texttt{#1}}
\expandafter\ifx\csname urlstyle\endcsname\relax
  \providecommand{\doi}[1]{doi: #1}\else
  \providecommand{\doi}{doi: \begingroup \urlstyle{rm}\Url}\fi

\bibitem[Adamkiewicz et~al.(2022)Adamkiewicz, Chen, Caccavale, Gardner, Culbertson, Bohg, and Schwager]{adamkiewicz2022vision}
Michal Adamkiewicz, Timothy Chen, Adam Caccavale, Rachel Gardner, Preston Culbertson, Jeannette Bohg, and Mac Schwager.
\newblock Vision-only robot navigation in a neural radiance world.
\newblock \emph{IEEE Robotics and Automation Letters}, 7\penalty0 (2):\penalty0 4606--4613, 2022.

\bibitem[Azinovi{\'c} et~al.(2022)Azinovi{\'c}, Martin-Brualla, Goldman, Nie{\ss}ner, and Thies]{azinovic2022neural}
Dejan Azinovi{\'c}, Ricardo Martin-Brualla, Dan~B Goldman, Matthias Nie{\ss}ner, and Justus Thies.
\newblock Neural rgb-d surface reconstruction.
\newblock In \emph{Proceedings of the IEEE/CVF Conference on Computer Vision and Pattern Recognition}, pages 6290--6301, 2022.

\bibitem[Betts and Huffman(1993)]{betts1993path}
John~T Betts and William~P Huffman.
\newblock Path-constrained trajectory optimization using sparse sequential quadratic programming.
\newblock \emph{Journal of Guidance, Control, and Dynamics}, 16\penalty0 (1):\penalty0 59--68, 1993.

\bibitem[Bharadhwaj et~al.(2020)Bharadhwaj, Xie, and Shkurti]{bharadhwaj2020model}
Homanga Bharadhwaj, Kevin Xie, and Florian Shkurti.
\newblock Model-predictive control via cross-entropy and gradient-based optimization.
\newblock In \emph{Learning for Dynamics and Control}, pages 277--286. PMLR, 2020.

\bibitem[Bialkowski et~al.(2011)Bialkowski, Karaman, and Frazzoli]{bialkowski2011massively}
Joshua Bialkowski, Sertac Karaman, and Emilio Frazzoli.
\newblock Massively parallelizing the rrt and the rrt.
\newblock In \emph{2011 IEEE/RSJ International Conference on Intelligent Robots and Systems}, pages 3513--3518. IEEE, 2011.

\bibitem[Bialkowski et~al.(2016)Bialkowski, Otte, Karaman, and Frazzoli]{bialkowski2016efficient}
Joshua Bialkowski, Michael Otte, Sertac Karaman, and Emilio Frazzoli.
\newblock Efficient collision checking in sampling-based motion planning via safety certificates.
\newblock \emph{The International Journal of Robotics Research}, 35\penalty0 (7):\penalty0 767--796, 2016.

\bibitem[Bohlin and Kavraki(2000)]{bohlin2000path}
Robert Bohlin and Lydia~E Kavraki.
\newblock Path planning using lazy prm.
\newblock In \emph{Proceedings 2000 ICRA. Millennium conference. IEEE international conference on robotics and automation. Symposia proceedings (Cat. No. 00CH37065)}, volume~1, pages 521--528. IEEE, 2000.

\bibitem[Camps et~al.(2022)Camps, Dyro, Pavone, and Schwager]{camps2022learning}
Gadiel~Sznaier Camps, Robert Dyro, Marco Pavone, and Mac Schwager.
\newblock Learning deep sdf maps online for robot navigation and exploration.
\newblock \emph{arXiv preprint arXiv:2207.10782}, 2022.

\bibitem[Chaplot et~al.(2021)Chaplot, Pathak, and Malik]{chaplot2021differentiable}
Devendra~Singh Chaplot, Deepak Pathak, and Jitendra Malik.
\newblock Differentiable spatial planning using transformers.
\newblock In \emph{International Conference on Machine Learning}, pages 1484--1495. PMLR, 2021.

\bibitem[Chen et~al.(2016)Chen, Liu, Liu, Miller, and How]{chen2016motion}
Yu~Fan Chen, Shih-Yuan Liu, Miao Liu, Justin Miller, and Jonathan~P How.
\newblock Motion planning with diffusion maps.
\newblock In \emph{2016 IEEE/RSJ International Conference on Intelligent Robots and Systems (IROS)}, pages 1423--1430. IEEE, 2016.

\bibitem[Chen and Zhang(2019)]{chen2019learning}
Zhiqin Chen and Hao Zhang.
\newblock Learning implicit fields for generative shape modeling.
\newblock In \emph{Proceedings of the IEEE/CVF Conference on Computer Vision and Pattern Recognition}, pages 5939--5948, 2019.

\bibitem[Choset et~al.(2005)Choset, Lynch, Hutchinson, Kantor, and Burgard]{choset2005principles}
Howie Choset, Kevin~M Lynch, Seth Hutchinson, George~A Kantor, and Wolfram Burgard.
\newblock \emph{Principles of robot motion: theory, algorithms, and implementations}.
\newblock MIT press, 2005.

\bibitem[Coifman and Lafon(2006)]{coifman2006diffusion}
Ronald~R Coifman and St{\'e}phane Lafon.
\newblock Diffusion maps.
\newblock \emph{Applied and computational harmonic analysis}, 21\penalty0 (1):\penalty0 5--30, 2006.

\bibitem[Connolly et~al.(1990)Connolly, Burns, and Weiss]{connolly1990path}
Christopher~I Connolly, J~Brian Burns, and Rich Weiss.
\newblock Path planning using laplace's equation.
\newblock In \emph{Proceedings., IEEE International Conference on Robotics and Automation}, pages 2102--2106. IEEE, 1990.

\bibitem[Crandall and Lions(1983)]{crandall1983viscosity}
Michael~G Crandall and Pierre-Louis Lions.
\newblock Viscosity solutions of hamilton-jacobi equations.
\newblock \emph{Transactions of the American mathematical society}, 277\penalty0 (1):\penalty0 1--42, 1983.

\bibitem[Crane et~al.(2013)Crane, Weischedel, and Wardetzky]{crane2013geodesics}
Keenan Crane, Clarisse Weischedel, and Max Wardetzky.
\newblock Geodesics in heat: A new approach to computing distance based on heat flow.
\newblock \emph{ACM Transactions on Graphics (TOG)}, 32\penalty0 (5):\penalty0 1--11, 2013.

\bibitem[Crane et~al.(2020)Crane, Livesu, Puppo, and Qin]{crane2020survey}
Keenan Crane, Marco Livesu, Enrico Puppo, and Yipeng Qin.
\newblock A survey of algorithms for geodesic paths and distances.
\newblock \emph{arXiv preprint arXiv:2007.10430}, 2020.

\bibitem[Dang et~al.(2020)Dang, Tranzatto, Khattak, Mascarich, Alexis, and Hutter]{dang2020graph}
Tung Dang, Marco Tranzatto, Shehryar Khattak, Frank Mascarich, Kostas Alexis, and Marco Hutter.
\newblock Graph-based subterranean exploration path planning using aerial and legged robots.
\newblock \emph{Journal of Field Robotics}, 37\penalty0 (8):\penalty0 1363--1388, 2020.

\bibitem[Fishman et~al.(2023)Fishman, Murali, Eppner, Peele, Boots, and Fox]{fishman2023motion}
Adam Fishman, Adithyavairavan Murali, Clemens Eppner, Bryan Peele, Byron Boots, and Dieter Fox.
\newblock Motion policy networks.
\newblock In \emph{Conference on Robot Learning}, pages 967--977. PMLR, 2023.

\bibitem[Gammell et~al.(2014)Gammell, Srinivasa, and Barfoot]{gammell2014informed}
Jonathan~D Gammell, Siddhartha~S Srinivasa, and Timothy~D Barfoot.
\newblock Informed {RRT}: Optimal sampling-based path planning focused via direct sampling of an admissible ellipsoidal heuristic.
\newblock In \emph{2014 IEEE/RSJ International Conference on Intelligent Robots and Systems}, pages 2997--3004. IEEE, 2014.

\bibitem[Han et~al.(2019)Han, Gao, Zhou, and Shen]{han2019fiesta}
Luxin Han, Fei Gao, Boyu Zhou, and Shaojie Shen.
\newblock Fiesta: Fast incremental euclidean distance fields for online motion planning of aerial robots.
\newblock In \emph{2019 IEEE/RSJ International Conference on Intelligent Robots and Systems (IROS)}, pages 4423--4430. IEEE, 2019.

\bibitem[Huh et~al.(2021)Huh, Isler, and Lee]{huh2021cost}
Jinwook Huh, Volkan Isler, and Daniel~D Lee.
\newblock Cost-to-go function generating networks for high dimensional motion planning.
\newblock In \emph{2021 IEEE International Conference on Robotics and Automation (ICRA)}, pages 8480--8486. IEEE, 2021.

\bibitem[Ichter et~al.(2018)Ichter, Harrison, and Pavone]{ichter2018learning}
Brian Ichter, James Harrison, and Marco Pavone.
\newblock Learning sampling distributions for robot motion planning.
\newblock In \emph{2018 IEEE International Conference on Robotics and Automation (ICRA)}, pages 7087--7094. IEEE, 2018.

\bibitem[Jaillet et~al.(2010)Jaillet, Cort{\'e}s, and Sim{\'e}on]{jaillet2010sampling}
L{\'e}onard Jaillet, Juan Cort{\'e}s, and Thierry Sim{\'e}on.
\newblock Sampling-based path planning on configuration-space costmaps.
\newblock \emph{IEEE Transactions on Robotics}, 26\penalty0 (4):\penalty0 635--646, 2010.

\bibitem[Kavraki et~al.(1996)Kavraki, Svestka, Latombe, and Overmars]{kavraki1996probabilistic}
Lydia~E Kavraki, Petr Svestka, J-C Latombe, and Mark~H Overmars.
\newblock Probabilistic roadmaps for path planning in high-dimensional configuration spaces.
\newblock \emph{IEEE transactions on Robotics and Automation}, 12\penalty0 (4):\penalty0 566--580, 1996.

\bibitem[Kavraki et~al.(1998)Kavraki, Kolountzakis, and Latombe]{kavraki1998analysis}
Lydia~E Kavraki, Mihail~N Kolountzakis, and J-C Latombe.
\newblock Analysis of probabilistic roadmaps for path planning.
\newblock \emph{IEEE Transactions on Robotics and automation}, 14\penalty0 (1):\penalty0 166--171, 1998.

\bibitem[Kuffner and LaValle(2000)]{kuffner2000rrt}
James~J Kuffner and Steven~M LaValle.
\newblock {RRT}-connect: An efficient approach to single-query path planning.
\newblock In \emph{Proceedings 2000 ICRA. Millennium Conference. IEEE International Conference on Robotics and Automation. Symposia Proceedings (Cat. No. 00CH37065)}, volume~2, pages 995--1001. IEEE, 2000.

\bibitem[Kumar et~al.(2019)Kumar, Mandalika, Choudhury, and Srinivasa]{kumar2019lego}
Rahul Kumar, Aditya Mandalika, Sanjiban Choudhury, and Siddhartha Srinivasa.
\newblock Lego: Leveraging experience in roadmap generation for sampling-based planning.
\newblock In \emph{2019 IEEE/RSJ International Conference on Intelligent Robots and Systems (IROS)}, pages 1488--1495. IEEE, 2019.

\bibitem[LaValle(2006)]{lavalle2006planning}
SM~LaValle.
\newblock Planning algorithms.
\newblock \emph{Cambridge University Press google schola}, 2:\penalty0 3671--3678, 2006.

\bibitem[Le et~al.(2023)Le, Chalvatzaki, Biess, and Peters]{le2023accelerating}
An~T. Le, Georgia Chalvatzaki, Armin Biess, and Jan Peters.
\newblock Accelerating motion planning via optimal transport.
\newblock In \emph{Advances in Neural Information Processing Systems (NeurIPS)}, 2023.

\bibitem[Li et~al.(2022{\natexlab{a}})Li, Xia, Mart\'in-Mart\'in, Lingelbach, Srivastava, Shen, Vainio, Gokmen, Dharan, Jain, Kurenkov, Liu, Gweon, Wu, Fei-Fei, and Savarese]{pmlr-v164-li22b}
Chengshu Li, Fei Xia, Roberto Mart\'in-Mart\'in, Michael Lingelbach, Sanjana Srivastava, Bokui Shen, Kent~Elliott Vainio, Cem Gokmen, Gokul Dharan, Tanish Jain, Andrey Kurenkov, Karen Liu, Hyowon Gweon, Jiajun Wu, Li~Fei-Fei, and Silvio Savarese.
\newblock igibson 2.0: Object-centric simulation for robot learning of everyday household tasks.
\newblock In Aleksandra Faust, David Hsu, and Gerhard Neumann, editors, \emph{Proceedings of the 5th Conference on Robot Learning}, volume 164 of \emph{Proceedings of Machine Learning Research}, pages 455--465. PMLR, 08--11 Nov 2022{\natexlab{a}}.
\newblock URL \url{https://proceedings.mlr.press/v164/li22b.html}.

\bibitem[Li et~al.(2022{\natexlab{b}})Li, Liu, Mello, Wang, Yang, and Kautz]{li2022learning}
Xueting Li, Sifei Liu, Shalini~De Mello, Xiaolong Wang, Ming-Hsuan Yang, and Jan Kautz.
\newblock Learning continuous environment fields via implicit functions.
\newblock In \emph{International Conference on Learning Representations}, 2022{\natexlab{b}}.
\newblock URL \url{https://openreview.net/forum?id=3ILxkQ7yElm}.

\bibitem[Li et~al.(2024)Li, Qiu, and Calinon]{li2024riemannian}
Yiming Li, Jiacheng Qiu, and Sylvain Calinon.
\newblock A riemannian take on distance fields and geodesic flows in robotics.
\newblock \emph{arXiv preprint arXiv:2412.05197}, 2024.

\bibitem[Lipman et~al.(2010)Lipman, Rustamov, and Funkhouser]{lipman2010biharmonic}
Yaron Lipman, Raif~M Rustamov, and Thomas~A Funkhouser.
\newblock Biharmonic distance.
\newblock \emph{ACM Transactions on Graphics (TOG)}, 29\penalty0 (3):\penalty0 1--11, 2010.

\bibitem[Loshchilov and Hutter(2017)]{Loshchilov2017DecoupledWD}
Ilya Loshchilov and Frank Hutter.
\newblock Decoupled weight decay regularization.
\newblock In \emph{International Conference on Learning Representations}, 2017.
\newblock URL \url{https://api.semanticscholar.org/CorpusID:53592270}.

\bibitem[Mescheder et~al.(2019)Mescheder, Oechsle, Niemeyer, Nowozin, and Geiger]{mescheder2019occupancy}
Lars Mescheder, Michael Oechsle, Michael Niemeyer, Sebastian Nowozin, and Andreas Geiger.
\newblock Occupancy networks: Learning 3d reconstruction in function space.
\newblock In \emph{Proceedings of the IEEE/CVF conference on computer vision and pattern recognition}, pages 4460--4470, 2019.

\bibitem[Millane et~al.(2023)Millane, Oleynikova, Wirbel, Steiner, Ramasamy, Tingdahl, and Siegwart]{Millane2023nvbloxGI}
Alexander Millane, Helen Oleynikova, Emilie Wirbel, Remo Steiner, Vikram Ramasamy, David Tingdahl, and Roland Siegwart.
\newblock nvblox: Gpu-accelerated incremental signed distance field mapping.
\newblock \emph{2024 IEEE International Conference on Robotics and Automation (ICRA)}, pages 2698--2705, 2023.
\newblock URL \url{https://api.semanticscholar.org/CorpusID:264832853}.

\bibitem[Newcombe et~al.(2011)Newcombe, Izadi, Hilliges, Molyneaux, Kim, Davison, Kohi, Shotton, Hodges, and Fitzgibbon]{newcombe2011kinectfusion}
Richard~A Newcombe, Shahram Izadi, Otmar Hilliges, David Molyneaux, David Kim, Andrew~J Davison, Pushmeet Kohi, Jamie Shotton, Steve Hodges, and Andrew Fitzgibbon.
\newblock Kinectfusion: Real-time dense surface mapping and tracking.
\newblock In \emph{2011 10th IEEE international symposium on mixed and augmented reality}, pages 127--136. Ieee, 2011.

\bibitem[Newcombe et~al.(2015)Newcombe, Fox, and Seitz]{newcombe2015dynamicfusion}
Richard~A Newcombe, Dieter Fox, and Steven~M Seitz.
\newblock Dynamicfusion: Reconstruction and tracking of non-rigid scenes in real-time.
\newblock In \emph{Proceedings of the IEEE conference on computer vision and pattern recognition}, pages 343--352, 2015.

\bibitem[Ni and Qureshi(2023{\natexlab{a}})]{Ni-RSS-23}
Ruiqi Ni and Ahmed~H Qureshi.
\newblock {Progressive Learning for Physics-informed Neural Motion Planning}.
\newblock In \emph{Proceedings of Robotics: Science and Systems}, Daegu, Republic of Korea, July 2023{\natexlab{a}}.
\newblock \doi{10.15607/RSS.2023.XIX.063}.

\bibitem[Ni and Qureshi(2023{\natexlab{b}})]{ni2023ntfields}
Ruiqi Ni and Ahmed~H Qureshi.
\newblock {NTF}ields: Neural time fields for physics-informed robot motion planning.
\newblock In \emph{International Conference on Learning Representations}, 2023{\natexlab{b}}.
\newblock URL \url{https://openreview.net/forum?id=ApF0dmi1_9K}.

\bibitem[Ni and Qureshi(2024)]{10610883}
Ruiqi Ni and Ahmed~H. Qureshi.
\newblock Physics-informed neural motion planning on constraint manifolds.
\newblock In \emph{2024 IEEE International Conference on Robotics and Automation (ICRA)}, pages 12179--12185, 2024.
\newblock \doi{10.1109/ICRA57147.2024.10610883}.

\bibitem[Ni et~al.(2021)Ni, Pan, and Gao]{Ni2021RobustMT}
Ruiqi Ni, Zherong Pan, and Xifeng Gao.
\newblock Robust multi-robot trajectory optimization using alternating direction method of multiplier.
\newblock \emph{IEEE Robotics and Automation Letters}, 7:\penalty0 5950--5957, 2021.
\newblock URL \url{https://api.semanticscholar.org/CorpusID:246634449}.

\bibitem[Oleynikova et~al.(2017)Oleynikova, Taylor, Fehr, Siegwart, and Nieto]{oleynikova2017voxblox}
Helen Oleynikova, Zachary Taylor, Marius Fehr, Roland Siegwart, and Juan Nieto.
\newblock Voxblox: Incremental 3d euclidean signed distance fields for on-board mav planning.
\newblock In \emph{2017 IEEE/RSJ International Conference on Intelligent Robots and Systems (IROS)}, pages 1366--1373. IEEE, 2017.

\bibitem[Ortiz et~al.(2022)Ortiz, Clegg, Dong, Sucar, Novotny, Zollhoefer, and Mukadam]{Ortiz-RSS-22}
Joseph Ortiz, Alexander Clegg, Jing Dong, Edgar Sucar, David Novotny, Michael Zollhoefer, and Mustafa Mukadam.
\newblock {iSDF: Real-Time Neural Signed Distance Fields for Robot Perception}.
\newblock In \emph{Proceedings of Robotics: Science and Systems}, New York City, NY, USA, June 2022.
\newblock \doi{10.15607/RSS.2022.XVIII.012}.

\bibitem[Pan and Manocha(2010)]{pan2010gpu}
Jia Pan and Dinesh Manocha.
\newblock Gpu-based parallel collision detection for real-time motion planning.
\newblock In \emph{Algorithmic Foundations of Robotics IX: Selected Contributions of the Ninth International Workshop on the Algorithmic Foundations of Robotics}, pages 211--228. Springer, 2010.

\bibitem[Park et~al.(2019)Park, Florence, Straub, Newcombe, and Lovegrove]{park2019deepsdf}
Jeong~Joon Park, Peter Florence, Julian Straub, Richard Newcombe, and Steven Lovegrove.
\newblock Deepsdf: Learning continuous signed distance functions for shape representation.
\newblock In \emph{Proceedings of the IEEE/CVF conference on computer vision and pattern recognition}, pages 165--174, 2019.

\bibitem[Qureshi and Yip(2018)]{qureshi2018deeply}
Ahmed~H Qureshi and Michael~C Yip.
\newblock Deeply informed neural sampling for robot motion planning.
\newblock In \emph{2018 IEEE/RSJ International Conference on Intelligent Robots and Systems (IROS)}, pages 6582--6588. IEEE, 2018.

\bibitem[Qureshi et~al.(2019)Qureshi, Simeonov, Bency, and Yip]{qureshi2019motion}
Ahmed~H Qureshi, Anthony Simeonov, Mayur~J Bency, and Michael~C Yip.
\newblock Motion planning networks.
\newblock In \emph{2019 International Conference on Robotics and Automation (ICRA)}, pages 2118--2124. IEEE, 2019.

\bibitem[Qureshi and Ayaz(2016)]{qureshi2016potential}
Ahmed~Hussain Qureshi and Yasar Ayaz.
\newblock Potential functions based sampling heuristic for optimal path planning.
\newblock \emph{Autonomous Robots}, 40:\penalty0 1079--1093, 2016.

\bibitem[Qureshi et~al.(2020)Qureshi, Miao, Simeonov, and Yip]{qureshi2020motion}
Ahmed~Hussain Qureshi, Yinglong Miao, Anthony Simeonov, and Michael~C Yip.
\newblock Motion planning networks: Bridging the gap between learning-based and classical motion planners.
\newblock \emph{IEEE Transactions on Robotics}, 37\penalty0 (1):\penalty0 48--66, 2020.

\bibitem[Raissi et~al.(2019)Raissi, Perdikaris, and Karniadakis]{raissi2019physics}
Maziar Raissi, Paris Perdikaris, and George~E Karniadakis.
\newblock Physics-informed neural networks: A deep learning framework for solving forward and inverse problems involving nonlinear partial differential equations.
\newblock \emph{Journal of Computational physics}, 378:\penalty0 686--707, 2019.

\bibitem[Ramsey et~al.(2024)Ramsey, Kingston, Thomason, and Kavraki]{Ramsey-RSS-24}
Clayton Ramsey, Zachary Kingston, Wil Thomason, and Lydia~E Kavraki.
\newblock {Collision-Affording Point Trees: SIMD-Amenable Nearest Neighbors for Fast Motion Planning with Pointclouds}.
\newblock In \emph{Proceedings of Robotics: Science and Systems}, Delft, Netherlands, July 2024.
\newblock \doi{10.15607/RSS.2024.XX.038}.

\bibitem[Sandstr{\"o}m et~al.(2023)Sandstr{\"o}m, Li, Van~Gool, and Oswald]{sandstrom2023point}
Erik Sandstr{\"o}m, Yue Li, Luc Van~Gool, and Martin~R Oswald.
\newblock Point-slam: Dense neural point cloud-based slam.
\newblock In \emph{Proceedings of the IEEE/CVF International Conference on Computer Vision}, pages 18433--18444, 2023.

\bibitem[Schulman et~al.(2014)Schulman, Duan, Ho, Lee, Awwal, Bradlow, Pan, Patil, Goldberg, and Abbeel]{schulman2014motion}
John Schulman, Yan Duan, Jonathan Ho, Alex Lee, Ibrahim Awwal, Henry Bradlow, Jia Pan, Sachin Patil, Ken Goldberg, and Pieter Abbeel.
\newblock Motion planning with sequential convex optimization and convex collision checking.
\newblock \emph{The International Journal of Robotics Research}, 33\penalty0 (9):\penalty0 1251--1270, 2014.

\bibitem[Sethian(1996)]{sethian1996fast}
James~A Sethian.
\newblock A fast marching level set method for monotonically advancing fronts.
\newblock \emph{Proceedings of the National Academy of Sciences}, 93\penalty0 (4):\penalty0 1591--1595, 1996.

\bibitem[Shen et~al.(2024)Shen, Peng, Yang, Xu, Bao, Hu, and Cui]{shen2024pc}
Xujie Shen, Haocheng Peng, Zesong Yang, Juzhan Xu, Hujun Bao, Ruizhen Hu, and Zhaopeng Cui.
\newblock Pc-planner: Physics-constrained self-supervised learning for robust neural motion planning with shape-aware distance function.
\newblock In \emph{SIGGRAPH Asia 2024 Conference Papers}, pages 1--11, 2024.

\bibitem[Sitzmann et~al.(2020)Sitzmann, Martel, Bergman, Lindell, and Wetzstein]{sitzmann2019siren}
Vincent Sitzmann, Julien~N.P. Martel, Alexander~W. Bergman, David~B. Lindell, and Gordon Wetzstein.
\newblock Implicit neural representations with periodic activation functions.
\newblock In \emph{Proc. NeurIPS}, 2020.

\bibitem[Stachniss(2009)]{stachniss2009robotic}
Cyrill Stachniss.
\newblock \emph{Robotic mapping and exploration}, volume~55.
\newblock Springer, 2009.

\bibitem[Sucar et~al.(2021)Sucar, Liu, Ortiz, and Davison]{sucar2021imap}
Edgar Sucar, Shikun Liu, Joseph Ortiz, and Andrew~J Davison.
\newblock imap: Implicit mapping and positioning in real-time.
\newblock In \emph{Proceedings of the IEEE/CVF International Conference on Computer Vision}, pages 6229--6238, 2021.

\bibitem[Tahir et~al.(2018)Tahir, Qureshi, Ayaz, and Nawaz]{tahir2018potentially}
Zaid Tahir, Ahmed~H Qureshi, Yasar Ayaz, and Raheel Nawaz.
\newblock Potentially guided bidirectionalized {RRT}* for fast optimal path planning in cluttered environments.
\newblock \emph{Robotics and Autonomous Systems}, 108:\penalty0 13--27, 2018.

\bibitem[Tamar et~al.(2016)Tamar, Wu, Thomas, Levine, and Abbeel]{tamar2016value}
Aviv Tamar, Yi~Wu, Garrett Thomas, Sergey Levine, and Pieter Abbeel.
\newblock Value iteration networks.
\newblock \emph{Advances in neural information processing systems}, 29, 2016.

\bibitem[Tancik et~al.(2020)Tancik, Srinivasan, Mildenhall, Fridovich-Keil, Raghavan, Singhal, Ramamoorthi, Barron, and Ng]{tancik2020fourier}
Matthew Tancik, Pratul Srinivasan, Ben Mildenhall, Sara Fridovich-Keil, Nithin Raghavan, Utkarsh Singhal, Ravi Ramamoorthi, Jonathan Barron, and Ren Ng.
\newblock Fourier features let networks learn high frequency functions in low dimensional domains.
\newblock \emph{Advances in Neural Information Processing Systems}, 33:\penalty0 7537--7547, 2020.

\bibitem[Thomason et~al.(2024)Thomason, Kingston, and Kavraki]{thomason2024motions}
Wil Thomason, Zachary Kingston, and Lydia~E Kavraki.
\newblock Motions in microseconds via vectorized sampling-based planning.
\newblock In \emph{2024 IEEE International Conference on Robotics and Automation (ICRA)}, pages 8749--8756. IEEE, 2024.

\bibitem[Thrun(2002)]{thrun2002probabilistic}
Sebastian Thrun.
\newblock Probabilistic robotics.
\newblock \emph{Communications of the ACM}, 45\penalty0 (3):\penalty0 52--57, 2002.

\bibitem[Treister and Haber(2016)]{treister2016fast}
Eran Treister and Eldad Haber.
\newblock A fast marching algorithm for the factored eikonal equation.
\newblock \emph{Journal of Computational physics}, 324:\penalty0 210--225, 2016.

\bibitem[Valero-Gomez et~al.(2013)Valero-Gomez, Gomez, Garrido, and Moreno]{valero2013path}
Alberto Valero-Gomez, Javier~V Gomez, Santiago Garrido, and Luis Moreno.
\newblock The path to efficiency: Fast marching method for safer, more efficient mobile robot trajectories.
\newblock \emph{IEEE Robotics \& Automation Magazine}, 20\penalty0 (4):\penalty0 111--120, 2013.

\bibitem[Von~Stryk and Bulirsch(1992)]{von1992direct}
Oskar Von~Stryk and Roland Bulirsch.
\newblock Direct and indirect methods for trajectory optimization.
\newblock \emph{Annals of operations research}, 37:\penalty0 357--373, 1992.

\bibitem[Williams et~al.(2016)Williams, Drews, Goldfain, Rehg, and Theodorou]{williams2016aggressive}
Grady Williams, Paul Drews, Brian Goldfain, James~M Rehg, and Evangelos~A Theodorou.
\newblock Aggressive driving with model predictive path integral control.
\newblock In \emph{2016 IEEE International Conference on Robotics and Automation (ICRA)}, pages 1433--1440. IEEE, 2016.

\bibitem[Xie et~al.(2022)Xie, Takikawa, Saito, Litany, Yan, Khan, Tombari, Tompkin, Sitzmann, and Sridhar]{xie2022neural}
Yiheng Xie, Towaki Takikawa, Shunsuke Saito, Or~Litany, Shiqin Yan, Numair Khan, Federico Tombari, James Tompkin, Vincent Sitzmann, and Srinath Sridhar.
\newblock Neural fields in visual computing and beyond.
\newblock In \emph{Computer Graphics Forum}, volume~41, pages 641--676. Wiley Online Library, 2022.

\bibitem[Xu et~al.(2021)Xu, Deng, and Shimada]{xu2021autonomous}
Zhefan Xu, Di~Deng, and Kenji Shimada.
\newblock Autonomous uav exploration of dynamic environments via incremental sampling and probabilistic roadmap.
\newblock \emph{IEEE Robotics and Automation Letters}, 6\penalty0 (2):\penalty0 2729--2736, 2021.

\bibitem[Yan et~al.(2021)Yan, Tian, Shi, Guo, Wang, and Zha]{yan2021continual}
Zike Yan, Yuxin Tian, Xuesong Shi, Ping Guo, Peng Wang, and Hongbin Zha.
\newblock Continual neural mapping: Learning an implicit scene representation from sequential observations.
\newblock In \emph{Proceedings of the IEEE/CVF International Conference on Computer Vision}, pages 15782--15792, 2021.

\bibitem[Yang et~al.(2022)Yang, Cao, Zhu, Oh, and Zhang]{yang2022far}
Fan Yang, Chao Cao, Hongbiao Zhu, Jean Oh, and Ji~Zhang.
\newblock Far planner: Fast, attemptable route planner using dynamic visibility update.
\newblock In \emph{2022 ieee/rsj international conference on intelligent robots and systems (iros)}, pages 9--16. IEEE, 2022.

\bibitem[Zhu et~al.(2021)Zhu, Peng, Larsson, Xu, Bao, Cui, Oswald, and Pollefeys]{Zhu2021NICESLAMNI}
Zihan Zhu, Songyou Peng, Viktor Larsson, Weiwei Xu, Hujun Bao, Zhaopeng Cui, Martin~R. Oswald, and Marc Pollefeys.
\newblock Nice-slam: Neural implicit scalable encoding for slam.
\newblock \emph{2022 IEEE/CVF Conference on Computer Vision and Pattern Recognition (CVPR)}, pages 12776--12786, 2021.
\newblock URL \url{https://api.semanticscholar.org/CorpusID:245385791}.

\bibitem[Zucker et~al.(2013)Zucker, Ratliff, Dragan, Pivtoraiko, Klingensmith, Dellin, Bagnell, and Srinivasa]{zucker2013chomp}
Matt Zucker, Nathan Ratliff, Anca~D Dragan, Mihail Pivtoraiko, Matthew Klingensmith, Christopher~M Dellin, J~Andrew Bagnell, and Siddhartha~S Srinivasa.
\newblock Chomp: Covariant hamiltonian optimization for motion planning.
\newblock \emph{The International journal of robotics research}, 32\penalty0 (9-10):\penalty0 1164--1193, 2013.

\end{thebibliography}

\begin{IEEEbiography}[{\includegraphics
[width=1in,height=1.25in,clip,
keepaspectratio]{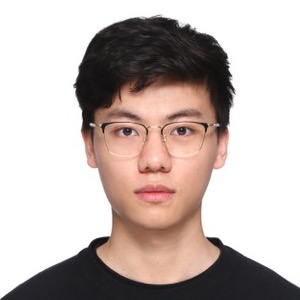}}]
{Yuchen Liu} (Student Member, IEEE) received his B.S. and M.S. degrees in computer science from New York University(NYU), New York, NY, USA, in 2022 and 2023, respectively. He is currently pursuing a Ph.D. in computer science at Purdue University, West Lafayette, IN, USA, where he is affiliated with the Cognitive Robot Autonomy and Learning Lab. His research interests span active mapping, motion planning, and physics-informed machine learning.
\end{IEEEbiography}

\begin{IEEEbiography}[{\includegraphics
[width=1in,height=1.25in,clip,
keepaspectratio]{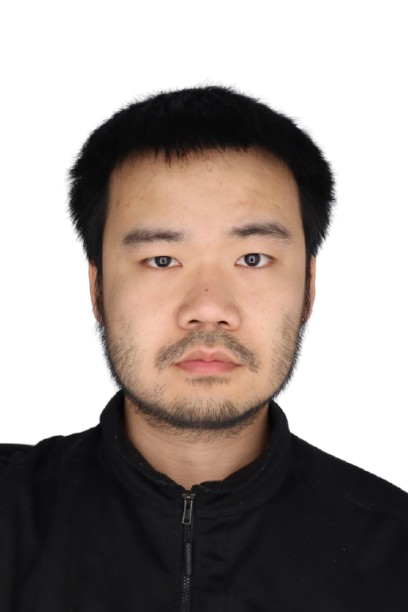}}]
{Ruiqi Ni} (Student Member, IEEE) received his B.S. degree in Information and Computational Science from the University of Science and Technology of China (USTC) in 2018. He later pursued graduate studies in Computer Science at Florida State University, FL, USA. Currently, he is a Ph.D. student in Computer Science at Purdue University and a member of the Cognitive Robot Autonomy and Learning Lab. His research focuses on motion planning and control, physics-informed machine learning, and physics-based simulation.

\end{IEEEbiography}

\begin{IEEEbiography}[{\includegraphics
[width=1.05in,height=1.5in,clip,
keepaspectratio]{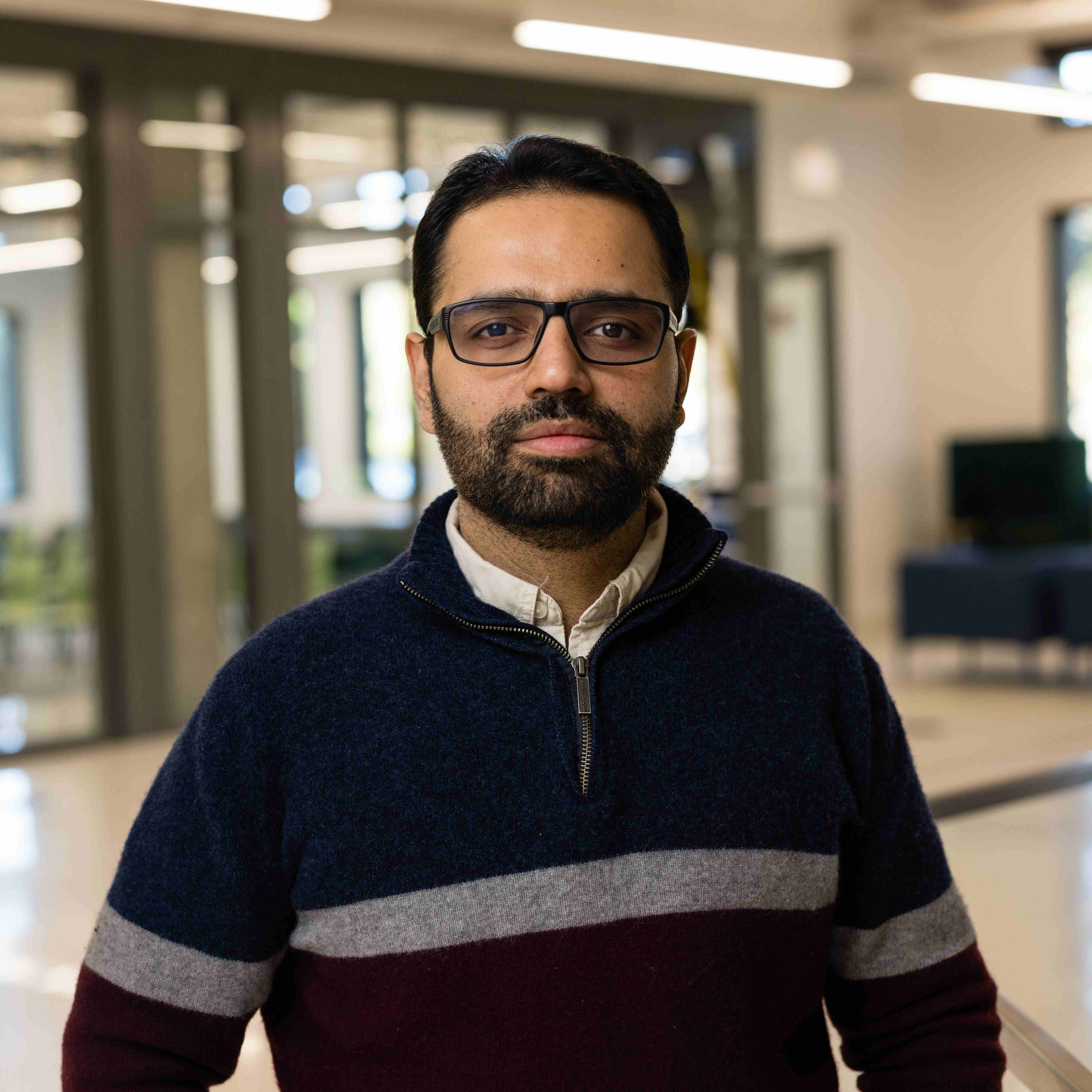}}] {Ahmed H. Qureshi} (Member, IEEE) is an Assistant Professor in the Department of Computer Science at Purdue University, where he leads the Cognitive Robot Autonomy and Learning (CoRAL) Lab. His research group focuses on both fundamental and applied aspects of robot planning and control, aiming to deploy robots in natural and dynamic human environments. His work addresses challenges such as scalable and rapid motion planning, active perception, human-in-the-loop robot manipulation, mobile navigation, and data-driven control. Dr. Qureshi's contributions to the field have been recognized through the spotlight and best paper awards at various academic venues. He currently serves as an Associate Editor for the IEEE Transactions on Robotics (TRO) and the IEEE Robotics and Automation Letters (RA-L). In 2024, he received the Outstanding Associate Editor Award from IEEE RA-L. He has also been involved in the program committees for prestigious conferences, including RSS, ICRA, IROS, and CoRL. Before his current position, Dr. Qureshi earned a B.S. in Electrical Engineering from NUST in Pakistan, an M.S. in Engineering from Osaka University in Japan, and a Ph.D. in Intelligent Systems, Robotics, and Control from the University of California, San Diego.

\end{IEEEbiography}

\end{document}